\documentclass[10pt,journal,compsoc]{IEEEtran}

\PassOptionsToPackage{table}{xcolor}

\ifCLASSOPTIONcompsoc
  \usepackage[nocompress]{cite}
\else
  \usepackage{cite}
\fi
\usepackage{subfloat}
\usepackage{subfig}
\usepackage{graphicx}
\usepackage{xcolor}
\usepackage{amsmath}
\usepackage[ruled,linesnumbered]{algorithm2e}
\usepackage{booktabs}
\usepackage{multirow}
\usepackage{makecell}
\usepackage{amssymb}
\usepackage{lipsum}
\usepackage{colortbl}
\usepackage{soul}
\usepackage{microtype}

\usepackage[colorlinks]{hyperref}

\definecolor{whitesmoke}{rgb}{0.96, 0.96, 0.96}

\newcommand{\revise}[1]{\textcolor{black}{#1}}
\newcommand{\major}[1]{\textcolor{black}{#1}}

\begin{document}
%\title{Spatial-Temporal Graph Attention Network for Multi-Frame 3D Object Detection}
% \title{Spatial-Temporal Enhanced Transformer Towards Multi-Frame 3D Object Detection}
\title{\major{Spatial-Temporal Graph Enhanced DETR Towards Multi-Frame 3D Object Detection}}
%缩写可以是: STEMD, STETR(3D), or STEP3D

%
%
% author names and IEEE memberships
% note positions of commas and nonbreaking spaces ( ~ ) LaTeX will not break
% a structure at a ~ so this keeps an author's name from being broken across
% two lines.
% use \thanks{} to gain access to the first footnote area
% a separate \thanks must be used for each paragraph as LaTeX2e's \thanks
% was not built to handle multiple paragraphs
%
%
%\IEEEcompsocitemizethanks is a special \thanks that produces the bulleted
% lists the Computer Society journals use for "first footnote" author
% affiliations. Use \IEEEcompsocthanksitem which works much like \item
% for each affiliation group. When not in compsoc mode,
% \IEEEcompsocitemizethanks becomes like \thanks and
% \IEEEcompsocthanksitem becomes a line break with idention. This
% facilitates dual compilation, although admittedly the differences in the
% desired content of \author between the different types of papers makes a
% one-size-fits-all approach a daunting prospect. For instance, compsoc 
% journal papers have the author affiliations above the "Manuscript
% received ..."  text while in non-compsoc journals this is reversed. Sigh.

\author{Yifan Zhang, Zhiyu Zhu, 
Junhui Hou,~\IEEEmembership{Senior Member, IEEE},  and Dapeng Wu,~\IEEEmembership{Fellow, IEEE}  
\IEEEcompsocitemizethanks{\IEEEcompsocthanksitem All authors are with the Department of Computer Science, City University of Hong Kong, Hong Kong. E-mail: yzhang3362-c@my.cityu.edu.hk; zhiyuzhu2-c@my.cityu.edu.hk; jh.hou@cityu.edu.hk; dapengwu@cityu.edu.hk.
\IEEEcompsocthanksitem This project was supported in part by the Hong Kong Research Grants Council
under Grant 11219422, Grant 11219324, and Grant 11218121, and in part by the Hong Kong Innovation and Technology Fund
under Grant MHP/117/21. \textit{Corresponding author: Junhui Hou} \protect \\
% note need leading \protect in front of \\ to get a newline within \thanks as
% \\ is fragile and will error, could use \hfil\break instead.

%\IEEEcompsocthanksitem XXXXXXX.
}% <-this % stops an unwanted space
%\thanks{Manuscript received April 19, 2005; revised August 26, 2015.}
}

% note the % following the last \IEEEmembership and also \thanks - 
% these prevent an unwanted space from occurring between the last author name
% and the end of the author line. i.e., if you had this:
% 
% \author{....lastname \thanks{...} \thanks{...} }
%                     ^------------^------------^----Do not want these spaces!
%
% a space would be appended to the last name and could cause every name on that
% line to be shifted left slightly. This is one of those "LaTeX things". For
% instance, "\textbf{A} \textbf{B}" will typeset as "A B" not "AB". To get
% "AB" then you have to do: "\textbf{A}\textbf{B}"
% \thanks is no different in this regard, so shield the last } of each \thanks
% that ends a line with a % and do not let a space in before the next \thanks.
% Spaces after \IEEEmembership other than the last one are OK (and needed) as
% you are supposed to have spaces between the names. For what it is worth,
% this is a minor point as most people would not even notice if the said evil
% space somehow managed to creep in.

% The paper headers
\markboth{}%
{Shell \MakeLowercase{\textit{et al.}}: Bare Demo of IEEEtran.cls for Computer Society Journals}
% The only time the second header will appear is for the odd numbered pages
% after the title page when using the twoside option.
% 
% *** Note that you probably will NOT want to include the author's ***
% *** name in the headers of peer review papers.                   ***
% You can use \ifCLASSOPTIONpeerreview for conditional compilation here if
% you desire.

% The publisher's ID mark at the bottom of the page is less important with
% Computer Society journal papers as those publications place the marks
% outside of the main text columns and, therefore, unlike regular IEEE
% journals, the available text space is not reduced by their presence.
% If you want to put a publisher's ID mark on the page you can do it like
% this:
%\IEEEpubid{0000--0000/00\$00.00~\copyright~2015 IEEE}
% or like this to get the Computer Society new two part style.
%\IEEEpubid{\makebox[\columnwidth]{\hfill 0000--0000/00/\$00.00~\copyright~2015 IEEE}%
%\hspace{\columnsep}\makebox[\columnwidth]{Published by the IEEE Computer Society\hfill}}
% Remember, if you use this you must call \IEEEpubidadjcol in the second
% column for its text to clear the IEEEpubid mark (Computer Society jorunal
% papers don't need this extra clearance.)

% use for special paper notices
%\IEEEspecialpapernotice{(Invited Paper)}

% for Computer Society papers, we must declare the abstract and index terms
% PRIOR to the title within the \IEEEtitleabstractindextext IEEEtran
% command as these need to go into the title area created by \maketitle.
% As a general rule, do not put math, special symbols or citations
% in the abstract or keywords.
\IEEEtitleabstractindextext{%
\begin{abstract}
The Detection Transformer (DETR) has revolutionized the design of CNN-based object detection systems, showcasing impressive performance. However, its potential in the domain of multi-frame 3D object detection remains largely unexplored. 
\major{In this paper, we present STEMD, a novel end-to-end framework that enhances the DETR-like paradigm for multi-frame 3D object detection by addressing three key aspects specifically tailored for this task.}
First, to model the inter-object spatial interaction and complex temporal dependencies, we introduce the spatial-temporal graph attention network, which represents queries as nodes in a graph and enables effective modeling of object interactions within a social context.
To solve the problem of missing hard cases in the proposed output of the encoder in the current frame, we incorporate the output of the previous frame to initialize the query input of the decoder.
Finally, it poses a challenge for the network to distinguish between the positive query and other highly similar queries that are not the best match. And similar queries are insufficiently suppressed and turn into redundant prediction boxes. To address this issue, our proposed IoU regularization term encourages similar queries to be distinct during the refinement.
Through extensive experiments, we demonstrate the effectiveness of our approach in handling challenging scenarios, while incurring only a minor additional computational overhead.
The code is publicly available at \url{https://github.com/Eaphan/STEMD}.
\end{abstract}

% Note that keywords are not normally used for peer review papers.
\begin{IEEEkeywords}
%Computer Society, IEEE, IEEEtran, journal, \LaTeX, paper, template.
Multi-Frame 3D Object Detection, Transformer, Graph Attention Network, Point Cloud, Autonomous Driving.
\end{IEEEkeywords}}

% make the title area
\maketitle

% To allow for easy dual compilation without having to reenter the
% abstract/keywords data, the \IEEEtitleabstractindextext text will
% not be used in maketitle, but will appear (i.e., to be "transported")
% here as \IEEEdisplaynontitleabstractindextext when the compsoc 
% or transmag modes are not selected <OR> if conference mode is selected 
% - because all conference papers position the abstract like regular
% papers do.
\IEEEdisplaynontitleabstractindextext
% \IEEEdisplaynontitleabstractindextext has no effect when using
% compsoc or transmag under a non-conference mode.

% For peer review papers, you can put extra information on the cover
% page as needed:
% \ifCLASSOPTIONpeerreview
% \begin{center} \bfseries EDICS Category: 3-BBND \end{center}
% \fi
%
% For peerreview papers, this IEEEtran command inserts a page break and
% creates the second title. It will be ignored for other modes.
\IEEEpeerreviewmaketitle

\IEEEraisesectionheading{\section{Introduction}\label{sec:introduction}}
%\IEEEPARstart{T}{his} demo file is intended to serve as a ``starter file'' for IEEE Computer Society journal papers produced under \LaTeX\ using IEEEtran.cls version 1.8b and later. I wish you the best of success.

\IEEEPARstart{T}{hree}-dimensional (3D) object detection is one of the fundamental tasks in the computer vision community that aims to identify and localize the oriented 3D bounding boxes of objects in specific classes. 
It plays a critical role in broad applications, including autonomous driving, object manipulation, and augmented reality. 
Recent years have witnessed the emergence of a large number of deep learning-based single-frame 3D detectors\cite{shi2020pv,zhang2023glenet,zhang2023upidet} with the advent of large-scale datasets\cite{Sun_2020_CVPR,caesar2020nuscenes}.
Nonetheless, given the intricacies of traffic environments, including long distances and inter-object occlusion, the object information encapsulated within point clouds may be inevitably subject to distortions of potential sparsity or incompleteness. 
Consequently, these aspects typically engender a sub-optimal performance of the single-frame 3D detectors~\cite{yin2021center}.

\begin{figure}[t]
	\centering
	\includegraphics[width=0.49\textwidth]{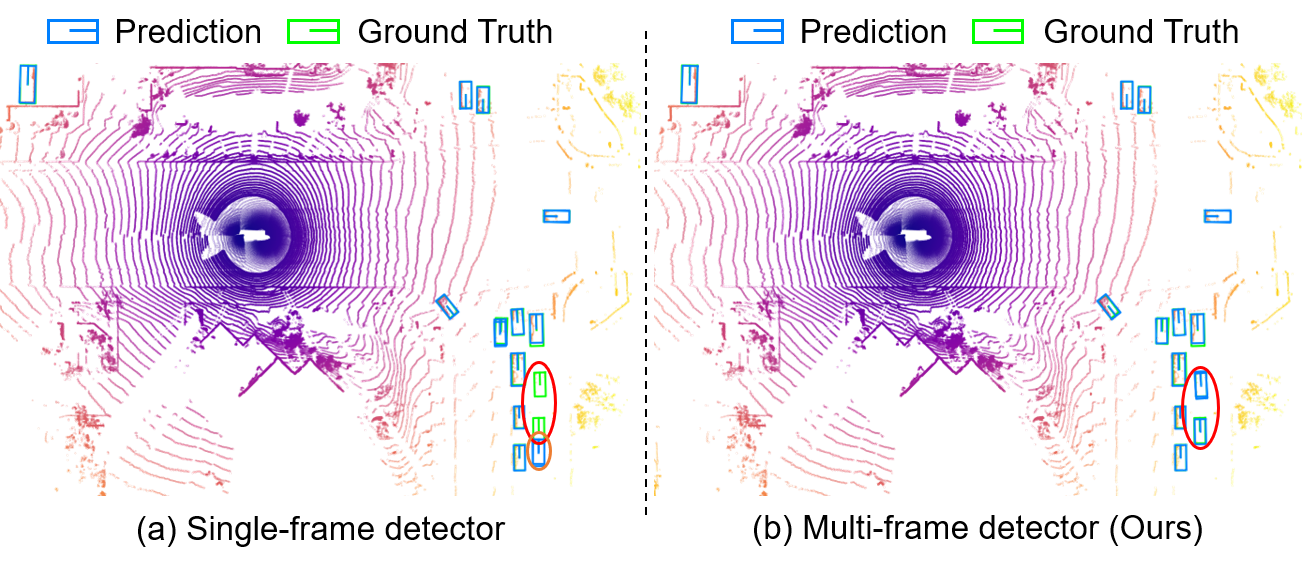} 
	\caption{
		\revise{Visual comparison of our multi-frame-based method and a single-frame detector~\cite{yin2021center} in a challenging outdoor 3D object detection scenario. As highlighted within \textcolor{red}{red} circles, our method leverages additional temporal information to successfully detect several heavily occluded objects, which the single-frame detector fails to detect. Moreover, there are more false-positive predictions in the result of the single-frame detector, as highlighted in \textcolor{orange}{orange} circles. Best viewed in color.}
	} \label{fig:challenge}\vspace{-0.4cm}
\end{figure}

As the point cloud sequence intrinsically provides multiple views of objects, it implies promising approaches to extract vital spatiotemporal information for facilitating more accurate detection, especially for objects that pose significant detection challenges.
By incorporating complementary information from other frames, a multi-frame 3D object detector exhibits improved performance compared to a single-frame 3D object detector (see Fig.~\ref{fig:challenge}).

The existing works in multi-frame 3D object detection have explored some feasible solutions.
A straightforward one is concatenating the observed points from multiple frames and using an additional dimension to indicate the timestamp~\cite{caesar2020nuscenes}. However, this method lacks explicit modeling of cross-frame relations and is less effective for fast-moving objects given multi-frame point cloud input~\cite{yang20213d}.
Some previous works naturally apply the long short-term memory (LSTM) network or gated recurrent unit (GRU) to voxel-level or BEV-level features for temporal modeling\cite{yin2021graph,huang2020lstm}.
3D-MAN~\cite{yang20213d} stores the features of box proposals in a memory bank and performs attention across proposal features from multiple perspectives. 
Recently some high-performance methods~\cite{chen2022mppnet,qi2021offboard,He_2023_CVPR} adopt a two-phase framework for multi-frame 3D object detection, where a baseline detector is employed to generate boxes and speed of objects in each frame, and the detected boxes across frames are associated to form a trajectory;
Then a specific region-based network is proposed to refine the box based on sequence object points and boxes.
Despite the gratifying success of these approaches, such two-phase pipelines for multi-frame object detection are somewhat sophisticated, requiring many extra hand-crafted components, e.g., IoU-based proposal matching, per-frame feature encoding, and trajectory-level feature propagation~\cite{chen2022mppnet}.
Thus, there is a pressing need to construct a novel framework for multi-frame 3D object detection that not only yields accurate results but also operates in a fully end-to-end fashion.

The recent advancements in sequential modeling~\cite{devlin2018bert} and cross-modal fusion~\cite{lu2019vilbert,prakash2021multi} reveal the remarkable capabilities of the Transformer architecture as a powerful framework for effectively modeling the information interaction within sequential or cross-modal data. The intrinsic self-attention module in Transformer plays a critical role in this success, as it enables the effective encoding of mutual relationships within the data. 
Especially, DETR~\cite{carion2020end} employs a transformer-based architecture to directly predict the bounding boxes and categories of objects, which replaces traditional convolutional neural network (CNN)-based detectors~\cite{yang20203dssd,ren2015faster} and achieves state-of-the-art performance~\cite{zhang2023dino}. 
The DETR paradigm is also well-suited for multi-frame 3D object detection but has not been well explored. 
Given the above factors, we propose \textbf{STEMD}, a novel end-to-end multi-frame 3D object detection framework based on the DETR-like paradigm in this paper. 
And we enhance the model from \textbf{\textit{three}} aspects specifically tailored for multi-frame 3D object detection, i.e., \textit{graph-based spatial-temporal modeling}, \textit{improved query initialization}, and an effective \textit{regularization term}, as detailed subsequently.

\begin{figure}[t]
	\centering
	\includegraphics[width=0.45\textwidth]{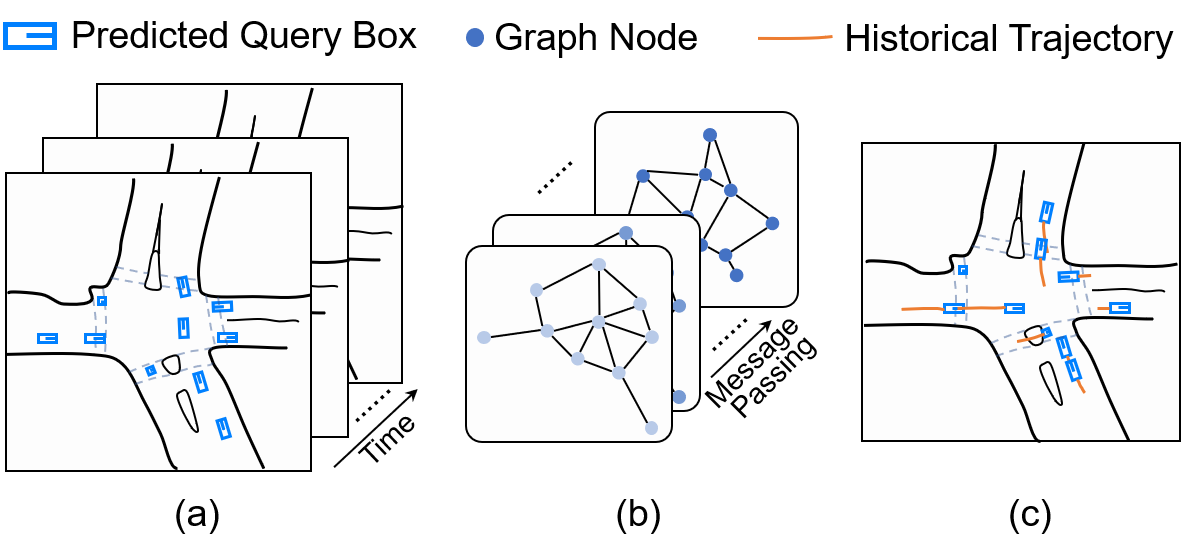}
	\caption{
		(a) Visualization of the multi-frame point clouds and the corresponding predicted query boxes within a scene.
		(b) Our approach represents the queries as nodes in a graph and utilizes the graph attention network (GAT) to effectively capture spatial-temporal dependencies.
		(c) The incorporated attention mechanism enables the model to learn interaction patterns among objects, empowering it to comprehend diverse social behaviors like collision avoidance, group following, adaptation to dynamic obstacles, and make precise predictions at current frame even in complex scenarios.
	} \label{fig:graph_motivation}
        % \vspace{-0.2cm}
\end{figure}

\textbf{1)} It has been widely acknowledged that scene understanding tasks, such as pedestrian trajectory prediction~\cite{li2022graph,yu2020spatio}, heavily rely on the effective modeling of spatial-temporal relationships between individual objects and their surroundings.
For example, in crowded environments, pedestrians exhibit diverse interaction patterns, such as avoiding collisions, following groups, or adapting to dynamic obstacles. 
However, in the context of multi-frame 3D object detection, we argue that the self-attention mechanism employed in the decoder of DETR fails to fully exploit the relations among queries. This limitation arises due to the dense application of self-attention across all queries.
To address this issue, we propose a graph-based attention network that leverages the complex spatial dependencies among objects. In our approach, we represent single objects (queries) as nodes and model their interactions as edges in a graph structure (as shown in Fig.~\ref{fig:graph_motivation}). By allowing nodes to dynamically attend to neighboring nodes in a context-aware manner, our graph-based attention network captures intricate interactions. 
\iftrue % comment for major revision
This attention mechanism facilitates the learning of various social behaviors, enabling our model to adapt to complex scenarios and make accurate predictions. Moreover, we introduce the graph-to-graph attention network to effectively model temporal dependencies. 
This enables our model to make current temporal predictions that satisfy the sequential consistency of object trajectories by attending to relevant past position information.
\fi

\textbf{2)} In existing DETR-like models~\cite{zhu2021deformable}, the encoder generates a set of region proposals, and the decoder progressively refines the bounding box predictions based on those queries initialized with these proposals. 
However, it is challenging to obtain accurate results for these cases through layer-by-layer refinement in the decoder if the initial queries inferred from the encoder miss some corner cases.
Since these cases may be more easily detectable in preceding frames, we propose to initialize additional input queries for the decoder in the present frame using selected final predicted boxes from the preceding frame.
Consequently, this initialization yields a higher recall rate of queries with respect to ground truth boxes and provides a strong starting point for refining the bounding box predictions.
This strategy, namely Temporal Query Recollection (TQR), helps in handling object occlusions, sudden disappearance, and other challenging scenarios commonly encountered in point cloud sequences.

\textbf{3)} \major{Existing DETR-based detectors~\cite{carion2020end,zhu2021deformable} utilize the one-to-one Hungarian Matching method, which assigns each ground truth box to the best matching query. It poses a challenge for the network to distinguish between the positive query and other highly similar queries that are not the best match. Since the DETR model does not utilize NMS (Non-Maximum Suppression), similar queries are insufficiently suppressed and turn into redundant prediction boxes.
To tackle these issues, we introduce an IoU regularization term that penalizes query boxes in close proximity to one another. This regularization encourages unmatched negative queries to differentiate their predicted bounding boxes from the positive queries, resulting in distinct queries and less redundant prediction boxes.}

We conduct extensive experiments on the prevailing Waymo dataset~\cite{Sun_2020_CVPR}. 
Our novel approach surpasses the performance of previous state-of-the-art single-frame and multi-frame 3D object detectors by a significant margin, while incurring only a tiny additional computation cost.

To summarize, the principal contributions of this research can be encapsulated as follows:
\begin{itemize}
    \item we propose to effectively model the socially-aware inter-object spatial interaction and complex temporal dependencies with a spatial-temporal graph attention network (STGA-Net), which represents queries as nodes in a graph;
    \item we propose TQR, a simple yet effective training strategy that enhances the initial query input of the decoder in the current frame using final predicted boxes from the previous frame; and
    \item we introduce an IoU regularization term to penalize query boxes with large overlaps, encouraging similar queries to output fewer redundant boxes as the ground-truth boxes in BEV generally do not overlap.
\end{itemize}

The remainder of the paper is organized as follows.
Sec. \ref{sec:related_work} reviews existing works most related to this work. 
In Sec.~\ref{sec:method}, we introduce the overall architecture of STEMD and elaborate on its key components. % of the proposed method.
In Sec.~\ref{sec:experiments}, we validate the effectiveness of our proposed method on the Waymo dataset and conduct ablation studies to analyze the effect of different components.
Finally, Sec.~\ref{sec:conclusion} concludes this paper.

\section{Related Work}\label{sec:related_work}
In this section, we mainly review existing works on single-frame 3D object detection, multi-frame 3D object detection, graph structure learning, and vision transformer, which are closely aligned with the core objectives of our work.\\

\noindent\textbf{Single-frame 3D Object Detection.}
Early research on single-frame 3D object detection can be classified into voxel-based and point-based approaches. Typically, voxel-based 3D detectors turn point clouds into grid-structure forms with fixed sizes and employ sparse convolution networks for feature extraction~\cite{Zhou_2018_CVPR, yan2018second, deng2021voxel}.
Point-based 3D detectors~\cite{shi2019pointrcnn, yang20203dssd} consume the raw 3D point clouds directly and extract highly semantic features through a series of downsampling and set abstraction layers following PointNet++~\cite{qi2017pointnet++}. 
There are also approaches that leverage a hybrid representation by integrating the multi-scale voxel-based features and point-based features containing accurate location information, and thus achieve a balance between detection accuracy and efficiency~\cite{shi2020pv, he2020structure}. PDV~\cite{hu2022point} efficiently localizes voxel features with voxel point centroids, which are then aggregated through a density-aware RoI grid pooling module using kernel density estimation and self-attention with point density positional encoding.\\

\noindent\textbf{Multi-frame 3D Object Detection.}
It has been proven in existing state-of-the-art works, given short point cloud sequence input, the simple concatenation of multi-frame point clouds can significantly outperform the single-frame detection~\cite{yin2021center,fan2022embracing}. However, this strategy lacks explicit modeling of cross-frame relations and is less effective for fast-moving objects given a long point cloud sequence~\cite{yang20213d,chen2022mppnet}.
\revise{Naturally, some early works applied LSTM or GRU to \textit{voxel-level} or \textit{BEV-level} feature maps across different frames for temporal modeling\cite{yin2021graph,huang2020lstm}. In contrast, our method models the \textit{object-level} spatial interaction and complex temporal dependencies.
Recently some high-performance methods~\cite{chen2022mppnet,qi2021offboard,He_2023_CVPR} adopted a \textit{two-phase} framework for multi-frame 3D object detection, where a baseline detector was employed to generate boxes and speed of objects in each frame, and the detected boxes across frames were associated to form a trajectory;
Then, a specific region-based network was proposed to refine the box based on sequence object points and boxes.
Despite the gratifying success of these approaches, such two-phase pipelines for multi-frame object detection are somewhat sophisticated, requiring many extra hand-crafted components, e.g., IoU-based proposal matching, per-frame feature encoding, and trajectory-level feature propagation~\cite{chen2022mppnet}. Our proposed STEMD is designed to be \textit{end-to-end} trainable and does not rely on additional supervision for the speed of objects.
\major{Unlike methods that rely on memory banks~\cite{yang20213d}, deformable attention~\cite{zhou2022centerformer}, or region-based networks~\cite{chen2022mppnet,He_2023_CVPR}, our STGA-Net uses a graph-based approach to model temporal interactions at the query level.}
}\\

%concatenate
%GNN (GRU), LSTM
%3d-man, centerformer, 
%MPPNet,
%Offboard
\noindent\textbf{Graph Structure Learning.}
\revise{
Graph neural networks (GNNs) have emerged as powerful tools for learning intrinsic representations of nodes and edges in graph-structured data \cite{wu2020comprehensive,kipf2016semi}. GNNs excel at capturing rich relationships among nodes and enabling comprehensive analysis of graph data. 
 GNNs have also been proven effective in various computer vision tasks, such as object detection~\cite{xu2019spatial}, pedestrian trajectory prediction~\cite{yu2020spatio,wang2020multi}, and modeling human body skeletons. 
Single-frame 3D detectors, including Point-GNN~\cite{shi2020point}, SVGA-Net~\cite{he2022svga}, PC-RGNN~\cite{zhang2021pc}, and Graph R-CNN~\cite{yang2022graph}, used GNNs to obtain stronger \textit{point-based representation} by modeling contextual information and mining point relations. Wang \textit{et al}.~\cite{wang2021semi} modeled the temporal relations between candidates in different time frames with a GNN to create more accurate detection results on unlabeled data for semi-supervised learning. In contrast, our work models the socially-aware inter-object spatial interaction and complex temporal dependencies between \textit{queries} in the decoder of Transformer with a spatial-temporal graph attention network. Although STEMD and these works employ graph structure learning, the motivation, operation, and applied scenarios are different.}  \\

\begin{figure*}[t]
	\centering
	\includegraphics[width=\textwidth]{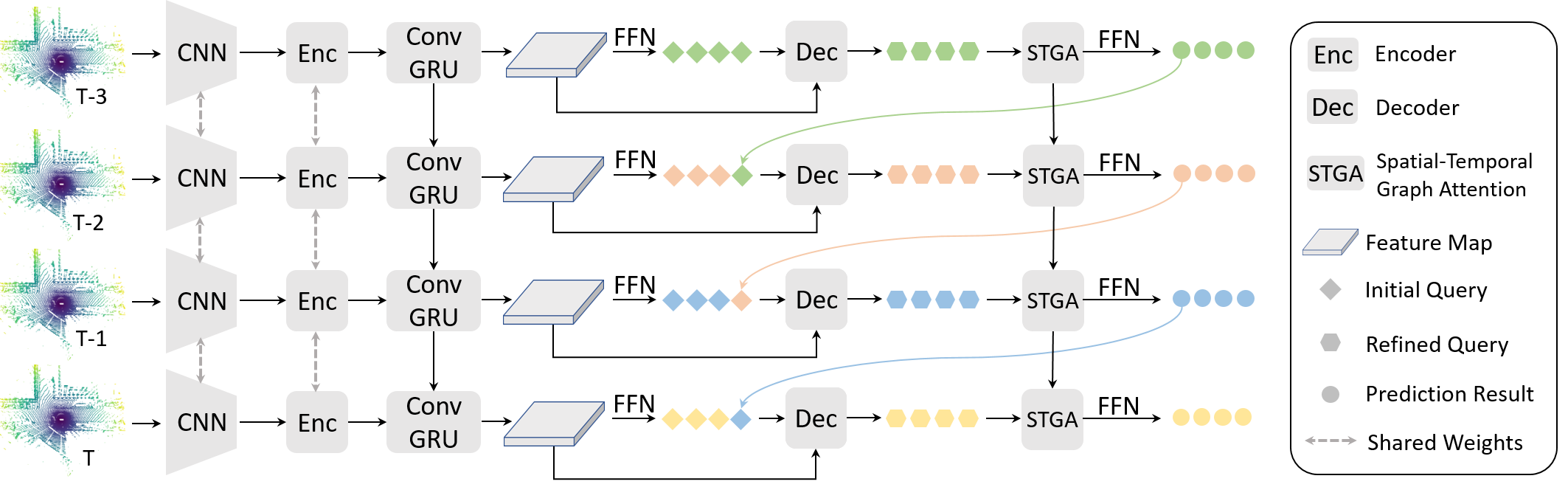} 
	\caption{
		%		Schematic of our multi-frame 3D object detection framework.
		The framework of the proposed STEMD. 
		The multi-frame point clouds are fed into the network as a sequence and a shared CNN extracts features of each frame. 
		Then the encoder performs local self-attention to enhance the features, followed by a ConvGRU block to model the feature-level temporal dependencies. 
		Next, a class-agnostic FFN is applied to generate object proposals, which are utilized to initialize the query input of the decoder.
		Subsequently, we use a graph attention network to capture both spatial and temporal dependencies among queries across different frames (see \S~\ref{sec:stga_net}).
		Especially, we select the high-scored final prediction of the last frame as extra query input of the decoder in the current frame, to mitigate the effect of the encoder missing hard cases in its predicted proposals (see \S~\ref{sec:tqr}).
		The STEMD could be trained in a fully end-to-end manner.
	} \label{fig:pipeline}
\end{figure*}

\noindent\textbf{Vision Transformer.}
Transformer-based models have gained significant popularity in recent years across various deep-learning tasks. Initially utilized in Natural Language Processing (NLP)~\cite{vaswani2017attention,devlin2018bert,radford2019language}, these models have witnessed success in computer vision tasks as well~\cite{dosovitskiy2020image,liu2021swin,carion2020end}. 
The intrinsic self-attention module in Transformer plays a critical role in this success, as it enables the effective encoding of mutual relationships within the data.
Especially, DETR presented a novel paradigm shift by formulating object detection as a direct set prediction problem~\cite{carion2020end}.
The core idea behind DETR lies in its ability to leverage the self-attention mechanism for capturing global context and modeling the relationships between objects in an image. Subsequently, Deformable DETR~\cite{zhu2021deformable} enhanced DETR by incorporating a deformable attention module that focuses on a small set of salient key elements in the feature map. 
Recently, some methods have also applied the transformer to 3D object detection tasks~\cite{zhou2022centerformer,zeng2022lift,sun2022swformer,yin2021graph,sheng2021improving}.
The ability of transformers to capture long-range dependencies and rich temporal information has made them a natural fit for multi-frame 3D object detection. By considering the motion and behavior of objects across frames, transformers can provide valuable insights into the dynamics of 3D objects.

\section{Proposed Method}\label{sec:method} % about 3.x~4 pages

\subsection{Overview}\label{sec:overview}
As illustrated in Figure~\ref{fig:pipeline}, architecturally, our framework follows the DETR-like paradigm and comprises several integral components, including a CNN-based backbone, an encoder, a decoder, and multiple FFN (Feed-Forward Network) prediction heads.
Furthermore, our framework incorporates convolutional GRU (ConvGRU) and spatial-temporal graph attention network, which are adopted to model feature-level and query-level spatial-temporal dependencies, respectively.
Specifically, our proposed model takes a long-term point cloud sequence of $T$ frames as input, denoted as  \{$I_1,..., I_T$\}. %as input that contains $T$ frames in total.
To eliminate the influence of ego-motion, we leverage the LiDAR pose information to align the point cloud of previous frames.
Then, point cloud $I_t$ in each time step $t$ is voxelized and forwarded to a sparse 3D convolution network to obtain the BEV features $X_t$. And we further perform local self-attention with BoxAttention~\cite{nguyen2022boxer} in the encoder module, followed by a ConvGRU block to model spatial-temporal dependencies and obtain enhanced encoder features $H_t$.
After that, a class-agnostic FFN is applied to generate object proposals based on the enhanced encoder features $H_t$, and the top $N_p$ scored proposals % $\mathcal{B}^E_t$ 
are selected to initialize the queries for the decoder.
In the decoder layers, we perform self-attention between queries and cross-attention between queries and the enhanced encoder features $H_t$~\cite{zhu2021deformable}. Here, the output of each decoder layer is then passed through a detection head to conduct iterative box refinement.
Particularly, we propose a spatial-temporal graph attention network to capture both the spatial relationship between queries in each frame and temporal dependencies between queries of adjacent frames. The enhanced graph embeddings are further passed to another detection head to obtain final predictions. 
Besides, 
to reduce the impact of the encoder missing hard cases in its proposal predictions, we select top $N_{\text{res}}$ scored prediction boxes of last frame to supplement these proposals as extra query input of decoder in current frame (see details in Sec.~\ref{sec:tqr}).

In what follows, we will introduce the key components in detail.

\begin{figure*}[t]
	\centering
	\includegraphics[width=\textwidth]{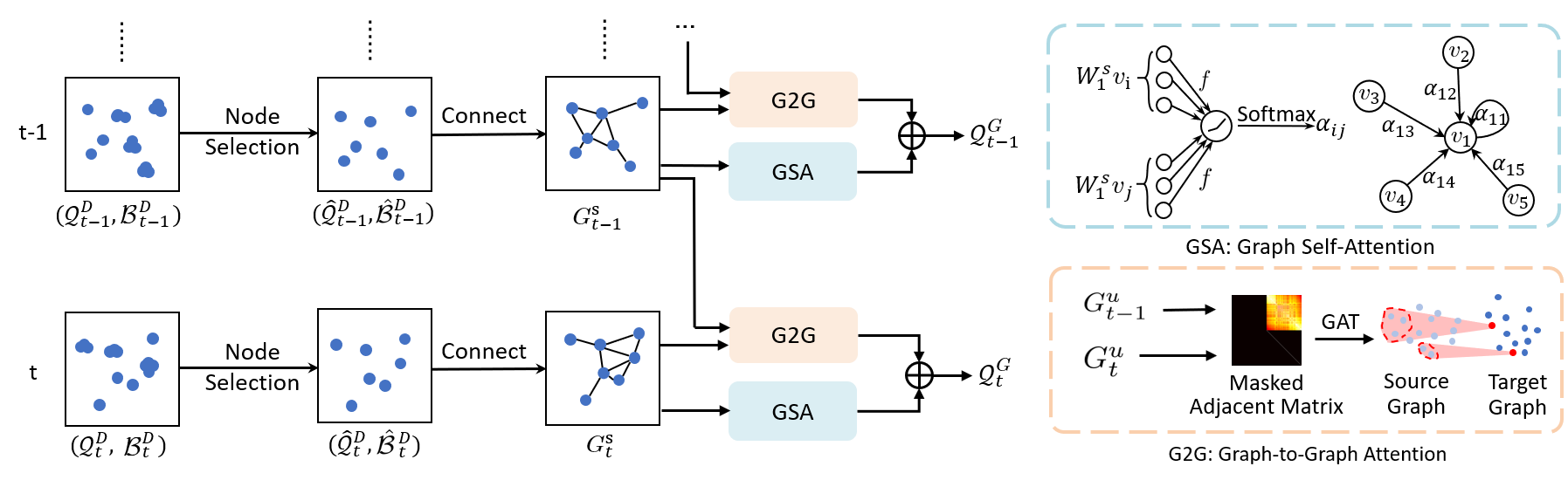} 
	\caption{
		\revise{Flowchart of the spatial-temporal graph attention network. First, we perform \textit{graph node selection} to suppress redundant query boxes and select representative ones from the candidate boxes as nodes in the graph. Then we use the graph self-attention to model the social interaction between objects. To effectively capture temporal dependencies across different frames, we incorporate the graph-to-graph attention network that enables message passing from the source graph (preceding frame) to the target graph (current frame). Then the spatial and temporal features are added to obtain new encodings for refining the queries.}
	} \label{fig:graph_attention}
\end{figure*}

\subsection{Spatial-Temporal Graph Attention Network}\label{sec:stga_net} % Graph Attention Learning % Spatial Temporal Transformer
Modeling spatial-temporal relationships between individual objects and their surroundings helps to understand complex scenarios and social interaction, thus yielding accurate detection results in the current frame.
As illustrated in Fig~\ref{fig:graph_attention}, we first introduce how to eliminate redundant bounding boxes output from the upstream decoder and derive representative queries for constructing graphs.
Next, we use the graph attention network to capture both \textit{spatial} relation between queries in a single frame and the \textit{temporal} dependencies between queries of adjacent frames.

\subsubsection{Graph Node Selection}\label{sec:graph_node_selection}
Redundant or overlapping bounding boxes predicted by the decoder can hinder the learning of graph structure topology and deteriorate the overall performance. To solve this problem, we propose an efficient method in this section, outlined in Algorithm~\ref{algorithm1}, to eliminate redundant bounding boxes and retain the filtered results as nodes in downstream graph attention modules. 
By doing so, we ensure that the selected bounding boxes effectively cover distinctive and representative regions of interest in the input data. 
This approach enhances the performance of graph-based learning and facilitates a more comprehensive understanding of the underlying relationships and structures within the scene.

Let $\mathcal{B}^{D}_{t}=\{b^{D}_{t,i}\}_{i=1}^{N_q}$, $\mathcal{S}^{D}_{t}=\{s^{D}_{t,i}\}_{i=1}^{N_q}$, and $\mathcal{Q}^{D}_{t}=\{q^{D}_{t,i}\}_{i=1}^{N_q}$ denote the predicted boxes, confidence scores, and query embeddings of the decoder at time $t$, respectively.
First, given a bounding box $b^{D}_{t,i}$, we define his neighboring sets $\mathcal{N}_{b^{D}_{t,i}}$, where $\texttt{IoU}(b^{D}_{t,i}, b^{D}_{t,j})>\theta$ for each  $j\in\mathcal{N}_{b^{D}_{t,i}}$. 
These neighboring sets contain boxes that overlap significantly with the given box. Second, we select the neighboring box $b^{D}_{t,j}$ with the highest confidence score, denoted as $b^{D}_{t,m}$ where $m$ is the index of the neighboring box with the highest confidence score.
Next, if the neighbors are not empty we update the confidence of $b^{D}_{t,i}$ as follows:

\begin{equation}
	\tilde{s}^{D}_{t,i} =
	\begin{cases}
		s^{D}_{t,i}, & \text{if } s^{D}_{t,i} \geq s^{D}_{t,m} \\
		s^{D}_{t,i} \times (1-\texttt{IoU}(b^{D}_{t,i}, b^{D}_{t,m})), & \text{if } s^{D}_{t,i} < s^{D}_{t,m}
	\end{cases}
\end{equation}
, where we suppress the scores of boxes that have more confident boxes around them~\cite{bodla2017soft}.
Finally, with the confidence scores of all boxes updated, we select bounding boxes with top $N_g$ scores.
The filtered bounding boxes $\mathcal{\hat{B}}^{D}_{t}=\{b^{D}_{t,i}\}_{i=1}^{N_g}$ and the corresponding query embedding $\mathcal{\hat{Q}}^{D}_{t}=\{q^{D}_{t,i}\}_{i=1}^{N_g}$ are sent to the subsequent spatial self-attention module and temporal cross-attention module for further processing.\\

\begin{algorithm}[t]
	% \normalem 
        %	\SetAlgoLined
        %	\SetAlgoNoLine
	\caption{Graph Node Selection}
	\label{algorithm1}
        %	\SetKwInOut{Input}{Input}
        %	\SetKwInOut{Output}{Output}
        %	\Input{
        %	\KwData{
		 \textbf{Input:} $\mathcal{B}^{D}_{t}=\{b^{D}_{t,i}\}_{i=1}^{N_q}$, $\mathcal{S}^{D}_{t}=\{s^{D}_{t,i}\}_{i=1}^{N_q}$, $\mathcal{Q}^{D}_{t}=\{q^{D}_{t,i}\}_{i=1}^{N_q}$ are the output bounding boxes, confidence scores, and queries of decoder.
%	}
	\BlankLine
	\For{ $i \gets 1$ \KwTo $N_q$ }{
		$\mathcal{N}_{b^{D}_{t,i}} = \{j \mid \texttt{IoU}(b^{D}_{t,i}, b^{D}_{t,j}) > \theta\}$\;
		$m = \underset{j}{\operatorname{argmax}} \left\{s^{D}_{t,j} \ \middle|\ j \in \mathcal{N}_{b^{D}_{t,i}}\right\}$\;
		\If{$\mathcal{N}_{b^{D}_{t,i}}$ is not empty}{
			\If{$s^{D}_{t,i} < s^{D}_{t,m}$}{
				$\tilde{s}^{D}_{t,i} \gets s^{D}_{t,i} \times (1-\texttt{IoU}(b^{D}_{t,i}, b^{D}_{t,m}))$\;
			}
		}
	}
	\BlankLine
	Select top $N_g$ bounding boxes based on confidence scores: $\mathcal{\hat{B}}^{D}_{t} \gets \{b^{D}_{t,i}\}_{i=1}^{N_g}$, $\mathcal{\hat{Q}}^{D}_{t} \gets \{q^{D}_{t,i}\}_{i=1}^{N_g}$\;
	\BlankLine
	\Return $\mathcal{\hat{B}}^{D}_{t}$, $\mathcal{\hat{Q}}^{D}_{t}$.
\end{algorithm}

\noindent \textbf{Discussion.} % Difference with NMS and soft-NMS. 
Unlike the NMS-series methods that rely on confidence ranking and are challenging to parallelize, our proposed method is highly parallelizable. Since each candidate box is only influenced by its neighboring bounding boxes, we can create $N_q$ threads to process them concurrently. This parallel processing significantly improves the efficiency of our algorithm.
Furthermore, our approach differs from traditional NMS-series methods in terms of the objective. 
The NMS post-processing improves the accuracy and efficiency of detectors by selecting the single best box and eliminating other duplicate bounding boxes, while our method aims to suppress redundant boxes and select representative ones from the candidate boxes as nodes in the graph and excessive inhibition may result in loss of essential node information.

\subsubsection{Spatial Self-Attention Module}

Given the selected output of the decoder for each time, we construct a spatial graph where each node corresponds to a query embedding. We define the spatial graph as $G_t^{s}=\{V_t^s,E_t^s\}$, where $V_t^s$ is the set of nodes and $E_t^s$ is the set of edges. 
Each node $v^{s}_{t,i}\in V_t^s$ corresponds to output query embedding $q^{D}_{t,i}$ of decoder. % and bounding box $b^{t,K}_{i}$. 
For each pair of nodes $(v^{s}_{t,i}, v^{s}_{t,j})$ in $V_t^s$, we calculate the distance between the centers of the corresponding query boxes $b^{D}_{t,i}$ and $b^{D}_{t,j}$.
If the distance between the two bounding boxes is below the defined threshold $d_s$, we add an edge between them to $E_t^s$. The edge indicates a spatial relationship or proximity between the corresponding query embeddings.

We use the graph attention network to capture spatial dependencies among the query embeddings in the spatial graph~\cite{velivckovic2017graph}. 
For a node $v^{s}_{t,i}$, we calculate the similarity coefficient between its neighbors  and itself:

\begin{equation}
%	e_{ij} = a([W_{1}^{s}q_i^{t,K} || W_{1}^{s}q_j^{t,K}]), j\in\mathcal{N}_{v_i^{t}}
	e_{ij} = f([W_{1}^{s}v^{s}_{t,i} || W_{1}^{s}v^{s}_{t,j}]),
\end{equation}
where $W_{1}^{s}$ is a learnable weight matrix, $||$ denotes concatenation $f$ is a single-layer feedforward neural network that maps the concatenated high-dimensional features to a real number, and $\mathcal{N}_{v^{s}_{t,i}}$ is the set of neighboring nodes of $v^{s}_{t,i}$ in the spatial graph.
Then we can further obtain the attention coefficients between each pair of nodes:
\begin{equation}
	\alpha_{ij} = \frac{\texttt{exp}(\texttt{LReLU}(e_{ij}))}{ {\textstyle \sum_{k'\in\mathcal{N}_{v^{s}_{t,i}}}}^{}\texttt{exp}(\texttt{LReLU}(e_{ik'}))},
\end{equation}
where $\texttt{LReLU}(\cdot)$ is a leaky rectified linear activation function.
The attention coefficients are used to compute a weighted sum of the feature vectors of neighboring nodes. Finally, the output of the spatial transformer is the sum of the original query embeddings $v^{s}_{t,i}$ and the computed feature vectors:

\begin{equation}
	\tilde{v}^{s}_{t,i} = v^{s}_{t,i} + \sigma(\sum_{j \in \mathcal{N}_{v^{s}_{t,i}}} \alpha_{ij}W_2^{s}v^{s}_{t,j}),
\end{equation}
where $W_2^{s}$ is a learnable weight matrix, $\sigma$ is a non-linear function.
Overall, the spatial self-attention modules model the interaction between the nodes, i.e., the query embedding in the same timestamp. 
%\zyf{refine the sentence?}

\subsubsection{Temporal Cross-Attention Module}
To capture temporal dependencies among the output of the decoder for multiple timestamps, we utilize the graph-to-graph attention mechanism.
Specifically, we use a GAT to model the temporal relationship between the source graph $G^u_{t-1}$ at $(t-1)$-th frame and the target graph $G^u_{t}$ at $t$-th frame. 
Let $V^u_t=\{v^{u}_{t,i}\}_{i=1}^{N_g}$ be the set of nodes in the target graph, where each node $v^{u}_{t,i}$ corresponds to the query embedding $q^{D}_{t,i}$. Similarly, we use $V^u_{t-1}=\{v^{u}_{t-1,i}\}_{i=1}^{N_g}$ to denote the set of nodes in the source graph.

%Specifically, for each node $v_i$ in the target graph, we compute the attention coefficients between $v_i$ and the nodes in the source graph:
The graph-to-graph cross-attention modules perform message passing from the source graph to the target graph. For the node $v^u_{t,i}$, we obtain its output by:

\begin{equation}
	%	v'^t_i = v^t_i + \sigma(\sum_{j \in \mathcal{N}_{v_i^t}} \beta_{ij}W_2^{s}q_j^{t,K}),
	\tilde{v}^{u}_{t,i} = v^{u}_{t,i} + \sigma(\sum_{j \in \mathcal{N}_{v^{u}_{t,i}}} \beta_{ij}W_1^{u}v^{u}_{t-1,j}),
\end{equation}
where $W^u_1$ is the learnable transformation weight matrix for the source graph, $\mathcal{N}_{v^{u}_{t,i}}$ is the neighboring nodes of $v^{u}_{t,i}$ in the source graph.
%Other symbols denote the same entities as in graph self-attention (Section 3.1.2). 
And $\beta_{ij}$ is the attention coefficient between $i$-th node of the source graph and $j$-th node of the target graph computed as:

\begin{equation}
	\beta_{ij} = \frac{\texttt{exp}(\texttt{LReLU}( f([W_{tgt}^{u}v^{u}_{t,i} || W_{src}^{u}v^{u}_{t-1,j}]) )) }
	{ {\textstyle \sum_{k'\in\mathcal{N}_{v^{u}_{t,i}}}}^{} \texttt{exp}(\texttt{LReLU}( f([W_{tgt}^{u}v^{u}_{t,i} || W_{src}^{u}v^{u}_{t-1,k'}]) )) },
\end{equation}
where $W_{src}^{u}$ and $W_{tgt}^{u}$ are the learnable transformation weight matrices for source and target graphs respectively.
Finally, we obtain the graph embedding by taking the sum of the output of the spatial self-attention module and temporal cross-attention module. 
This informative graph embedding is further passed to another prediction head, which we call the graph head, to achieve refined bounding boxes.
%In such wise, we are able to capture both spatial and temporal dependencies among the output of the decoder for multi frames, which helps to improve the accuracy of 3D object detection.

\begin{table}
	\centering
	\caption{
		\major{Comparison of the STGA-Net with conventional self-attention regarding time complexity.
			${\bar N}^s_E$ and ${\bar N}^U_E$ denote the average number of neighbors for nodes in the spatial graph and target graph, respectively. $C$ represents the dimension of the query embedding.}	
	}
	\label{table:stga_time_complexity}
	\begin{tabular}{lll} 
		\hline
		Method       & Phase      & Time Complexity  \\ 
		\hline
		\multirow{3}{*}{STGA-Net~} & Node Selection     & $\mathcal{O}(N_q^2)$  \\
		& Spatial Self-Attention   						& $\mathcal{O}(N_g{\bar N}^s_EC)$  \\
		& Temporal Cross-Attention 						& $\mathcal{O}(N_g{\bar N}^u_EC)$  \\ 
		\hline
		Self-Attention             & Total      		& $\mathcal{O}(N_q^2C)$  \\
		\hline
	\end{tabular}
\end{table}

\noindent \textbf{\major{Time Complexity Analysis}}. \major{The time complexity of the STGA-Net and the conventional self-attention mechanism is compared in Table~\ref{table:stga_time_complexity}. The graph node selection phase involves identifying and filtering redundant bounding boxes based on IoU thresholds. For each bounding box, identify neighbors with an IoU above a certain threshold, leading to a complexity of $\mathcal{O}(N_q^2)$, where $N_q$ is the number of initial query embeddings, and adjusting confidence scores involves sorting, which has a complexity of $\mathcal{O}(N_q\rm{log}N_q)$. Overall, the time complexity of graph node selection is dominated by the neighbor identification step, resulting in $\mathcal{O}(N_q^2)$.}

\major{In the spatial self-attention module, constructing the spatial graph involves connecting nodes based on a distance threshold. This step has a complexity of $\mathcal{O}(N_g^2)$, where $\mathcal{O}(N_g)$ is the number of nodes in the spatial graph. For each node, calculating attention weights involves operations with its ${\bar N}^s_E$ neighbors. Given $C$ as the dimensionality of the feature vectors, the complexity per node is $\mathcal{O}({\bar N}^s_EC)$. For all $N_g$ nodes, this results in $\mathcal{O}(N_g{\bar N}^s_EC)$. Overall, the time complexity of the spatial self-attention module is $\mathcal{O}(N_g{\bar N}^s_EC)$. Similarly, the time complexity of the temporal cross-attention module is $\mathcal{O}(N_g{\bar N}^u_EC)$, where ${\bar N}^u_E$ is the average number of neighbors for nodes in the target graph.}

\major{In comparison, the conventional self-attention mechanism treats each query as attending to all other queries. This results in a total time complexity of $\mathcal{O}(N_q^2C)$, where every query interacts with every other query, leading to a quadratic complexity in terms of the number of queries.}

\major{Overall, while the conventional self-attention has a simpler structure with a $\mathcal{O}(N_q^2C)$ complexity, the STGA-Net leverages structured interactions, breaking down the problem into manageable spatial and temporal components. This structured approach often leads to more efficient computations, especially when the average number of neighbors ${\bar N}^s_E$ and ${\bar N}^u_E$ is significantly smaller than the total number of queries $N_q$.}

\subsection{Temporal Query Recollection}\label{sec:tqr}
\noindent \textbf{Motivation}. Current models like DETR~\cite{zhu2021deformable} generate region proposals in the encoder, and the decoder progressively refines the bounding box predictions based on these queries initialized with the proposals. However, if the initial queries inferred from the encoder miss some corner cases, it is challenging to obtain accurate results for these cases through layer-by-layer refinement in the decoder. 
To address this limitation, we propose a training strategy called temporal query recollection to enhance the initial queries of the decoder. Specifically, we incorporate the final predictions from the last frame as an additional query input for the decoder in the current frame. The reason behind employing this strategy is that challenging cases that are missed by the encoder at the current frame might be comparatively easier to detect in the preceding frame.

Let $\mathcal{B}^{G}_{t-1}=\{b^{G}_{t-1,i}\}_{i=1}^{N_g}$ and $\mathcal{S}^{G}_{t-1}=\{s^{G}_{t-1,i}\}_{i=1}^{N_g}$ denote the prediction boxes and scores from the graph head at last timestamp $t-1$.
When initializing the query input for the decoder, we select top $N_{\text{res}}$ scored boxes $\hat{\mathcal{B}}^{G}_{t-1}$ as the supplementary of prediction of encoder $\mathcal{B}^{E}_t=\{b^{E}_{t,i}\}_{i=1}^{N_p}$.
The total number of initial query embeddings $N_q$ for the decoder is $N_q=N_p + N_{\text{res}}$.
%The $N_q=N_p + N_{\text{res}}$ initial query embeddings $\mathcal{Q}^{0}_t$ for decoder is derived by:
The initial query embeddings $\mathcal{Q}^{0}_t$ for the decoder are derived as follows:
\begin{equation}
	\mathcal{Q}^{0}_t=\{q^{0}_{t,i} = \text{PE}(b^{0}_{t,i}, s^{0}_{t,i}), i=1,...,N_q\},
\end{equation}
where each initial query embedding $q^{0}_{t,i}$ is encoded from the bounding box $b^{0}_{t,i}$ and confidence $s^{0}_{t,i}$ through the position encoding (PE).

Consequently, this initialization strategy improves the recall rate of initial queries with respect to the ground-truth boxes and provides a strong starting point for refining the bounding box predictions in the current frame. 
The TQR strategy helps handle challenging scenarios such as object occlusions, sudden disappearances, and other common difficulties encountered in 3D object detection.

\subsection{IoU Regularization Term}

\begin{figure}[t]
	\centering
	\includegraphics[width=0.48\textwidth]{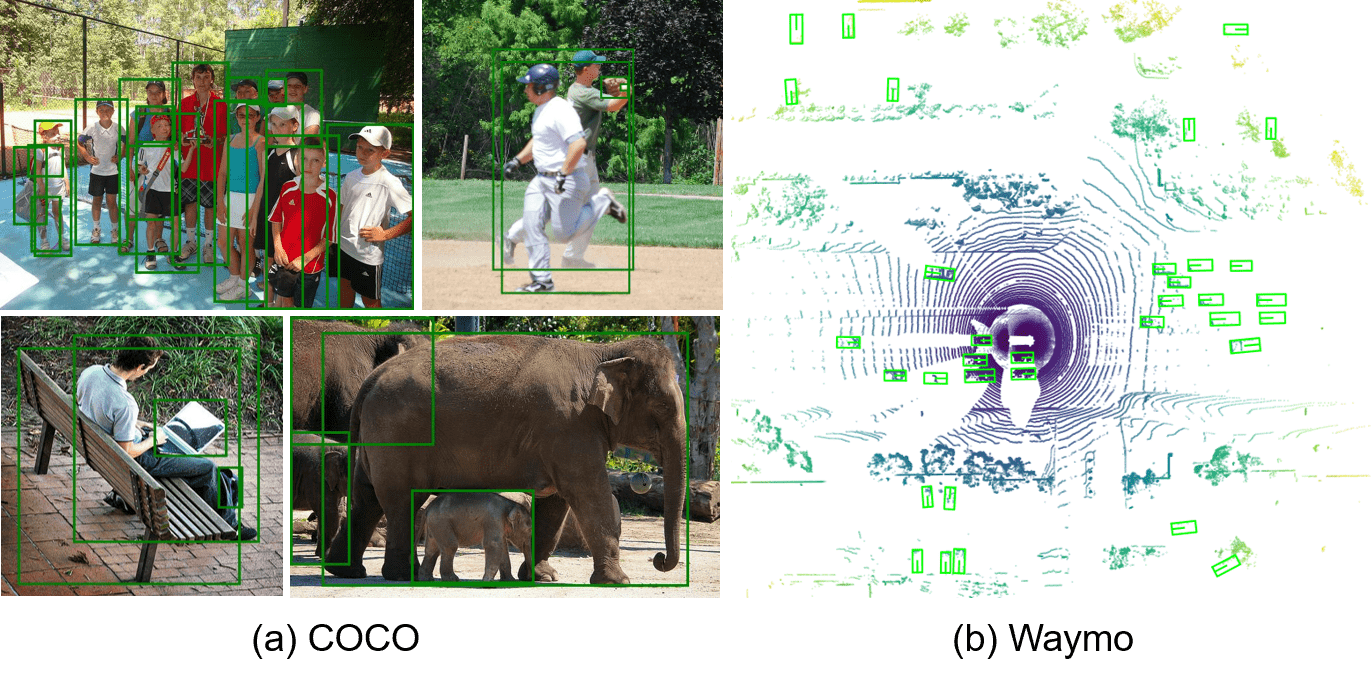} 
	\caption{
		Samples of the COCO dataset (left) and the Waymo dataset (right). The green boxes denote the ground truth bounding boxes. Unlike 2D detection, the bounding boxes in the Bird's Eye View generally do not overlap in the 3D object detection task.
	} 
	\label{fig:dataset}
\end{figure}

\major{DETR-like detectors generally use a one-to-one Hungarian Matching method. In this method, each ground truth box is assigned only the best matching query based on the classification cost and bounding box regression cost. Queries that are close to the ground truth box but have a larger cost are considered negative samples. It poses a challenge for the network to optimize as similar queries are assigned opposite labels under one-to-one assignment. If we set the number of queries to be small, the sparse queries fall short in recall rate. However, the current detection loss supervision could not solve such a dilemma well. Furthermore, since the DETR model does not utilize NMS (Non-Maximum Suppression), similar queries are insufficiently suppressed and turn into redundant prediction boxes.}

\major{To address this issue, we introduce an IoU regularization term that encourages similar queries to differentiate during refinement. This regularization term penalizes query boxes that are in close proximity to each other. By doing so, it encourages the unmatched queries to adjust their predicted bounding boxes away from the matched query, even if they are highly similar. We provide the observation that, in 3D object detection tasks, unlike 2D detection, the bounding boxes in the Bird's Eye View generally do not overlap (See Fig.~\ref{fig:dataset}). Therefore, the IoU regularization term does not affect the relationship between the matched queries as they generally do not overlap.}

Specifically, we consider a specific FFN prediction head in our model that outputs a set of query boxes denoted as $\mathcal{B}=\{b_i\}_{i=1}^{N'}$. We introduce a regularization loss term, denoted as $\mathcal{R}_b$, which is added to the overall loss function during model optimization. This term is computed by summing the IoU between each pair of bounding boxes, weighted by their corresponding confidence scores:
\begin{equation}
	\mathcal{R}_b = \sum_{i=1}^{N'} \sum_{j=1,j\neq i}^{N'} s_i \times \texttt{IoU}(b_i, b_j),
\end{equation}
where $s_i$ is the corresponding confidence score of bounding box $b_i$. 
The introduction of a confidence score aims to place more emphasis on bounding boxes that are close to the ground truth object with a high confidence score but are not the best match. These boxes are pushed further away from the best match, while boxes with lower confidence scores are considered less important and assigned lower weights.

\subsection{Loss Function and Inference}\label{sec:loss}
\textbf{Loss function}. During training, we leverage the Hungarian algorithm to assign ground truths to the object queries in all FFN prediction heads in the encoder, decoder, and graph head. 
Following ~\cite{carion2020end,zhu2022conquer}, we define the loss of each FFN prediction head as:
\begin{equation}
	L = \lambda_{\rm{cls}} L_{\rm{cls}} + \lambda_{\rm{h}} L_{\rm{Huber}} + \lambda_{\rm{giou}} L_{\rm{GIoU}} + \lambda_{\rm{r}} \mathcal{R}_{b},
\end{equation}
where we adopt the focal loss $L_{cls}$ for classification, the Huber loss $L_{Huber}$ and 3D GIoU loss $L_{GIoU}$ are utilized for box regression,
$\lambda_{\rm{cls}}, \lambda_{\rm{h}}, \lambda_{\rm{giou}}, \lambda_{\rm{r}}$ are hyper-parameters to balance the penalty terms.\\ 

\noindent \textbf{Inference}. 
For online multi-frame 3D object detection, we do not repeat all steps %including encoder, decoder, etc. 
for the previous frames when we perform detection at the current frame.
Instead, we preserve useful variables during the detection of the last frame $t-1$. 
Specifically, when conducting detection at current time $t$, the hidden state $H_{t-1}$ is used in ConvGRU modules to perform feature-level temporal enhancement, and the query embeddings inferred by the decoder in last frame $t-1$ are preserved to serve as nodes $V^u_{t-1}$ in the source graph $G^u_{t-1}$ to perform spatial-temporal message passing to target graph $G^u_{t}$.
Therefore, the inference speed is not much slower than the single-frame detector.
%\zyf{More details of the XXX can be found in XXX.}

\section{Experiments}\label{sec:experiments}
In this section, we describe the datasets and evaluation metrics in Sec.~\ref{sec:dataset}, and introduce the implementation details in Sec.~\ref{sec:implementation_details}. Then we compare our method with previous state-of-the-art methods in Sec.~\ref{sec:waymo_val_result}. Afterward, thorough ablative studies are conducted to investigate the effectiveness of essential components of our framework in Sec.~\ref{sec:ablation_study}. Finally, we also provide the run-time efficiency analysis for online 3D object detection scenarios in Sec~\ref{sec:efficiency}.

\subsection{Dataset and Evaluation Metric}\label{sec:dataset}
\revise{
We evaluate our method on the \textit{Waymo Open Dataset}~\cite{Sun_2020_CVPR} and \textit{nuScenes} dataset~\cite{caesar2020nuscenes}.
% The Waymo Open Dataset is a comprehensive and diverse dataset specifically designed for 3D object detection in autonomous driving scenarios. 
The Waymo Open Dataset consists of a large collection of 798 sequences for training, 202 sequences for validation, and 150 sequences for testing, resulting in a total of 198,438 LiDAR frames. Each sequence provides a rich set of data, including LiDAR point clouds, multi-view camera images, and object annotations spanning a full 360-degree field of view.
Each sequence contains approximately 200 frames spanning 20 seconds. 
%The Waymo dataset employs a 64-beam LiDAR with a capture frequency of 10Hz, resulting in around 180,000 points per frame.
To facilitate the evaluation of 3D object detectors, the objects in the Waymo Open Dataset are categorized into two difficulty levels: LEVEL\_L1, which represents objects with more than five observed LiDAR points, and LEVEL\_L2, which includes objects with 1-5 points. The performance of 3D object detectors is commonly assessed using the mean Average Precision (mAP) and mean Average Precision weighted by heading accuracy (mAPH) metrics.
The nuScenes dataset comprises 700 training sequences and 150 validation sequences. Each sequence in this dataset spans approximately 20 seconds, featuring a frame every 0.05 seconds. Annotation is provided for keyframes, which are defined as every tenth frame in a sequence. The employed metrics in evaluation are the mean average precision (mAP) and the nuScenes detection score (NDS)~\cite{caesar2020nuscenes}.
}

\setlength{\tabcolsep}{5pt}
\begin{table*}
	\centering
	\caption{\major{Comparison with state-of-the-art methods on the validation set of Waymo Open Dataset (train with 100\% training data). ${}^\dagger$ indicates that methods use the two-phase training framework and depend on extra speed supervision. The best results are highlighted in \textbf{bold}.}}
	\label{table:waymo_val}
	\resizebox{\textwidth}{!}{
		\begin{tabular}{c|c|c|cc|cc|cc|cc} 
			\toprule
			\multirow{2}{*}{Methods} & \multirow{2}{*}{Present at} & \multirow{2}{*}{Frames} & \multicolumn{2}{c|}{mAP/mAPH} & \multicolumn{2}{c|}{Vehicle 3D AP/APH} & \multicolumn{2}{c|}{Pedestrian 3D AP/APH} & \multicolumn{2}{c}{Cyclist 3D AP/APH}  \\
			&  &  & L1 & L2  & L1 & L2  & L1 & L2 & L1 & L2  \\ 
			\midrule
            SECOND~\cite{yan2018second} & Sensors'18 & 1 & 67.2/63.1 & 61.0/57.2 & 72.3/71.7 & 63.9/63.3 & 68.7/58.2 & 60.7/51.3 & 60.6/59.3 & 58.3/57.0 \\
            % PointPillars~\cite{Lang_2019_CVPR} & CVPR'19 & 1 & 69.0/63.5 & 62.8/57.8 & 72.1/71.5 & 63.6/63.1 & 70.6/56.7 & 62.8/50.3 & 64.4/62.3 & 61.9/59.9 \\
            PV-RCNN~\cite{shi2020pv} & CVPR'20 & 1 & 76.2/73.6 & 69.6/67.2 & 78.0/77.5 & 69.4/69.0 & 79.2/73.0 & 70.4/64.7 & 71.5/70.3 & 69.0/67.8 \\
            Part-A2-Net~\cite{shi2020points} & TPAMI'20 & 1 & 73.6/70.3 & 66.9/63.8 & 77.1/76.5 & 68.5/68.0 & 75.2/66.9 & 66.2/58.6 & 68.6/67.4 & 66.1/64.9 \\
            LiDAR-RCNN~\cite{li2021lidar} & CVPR'21 & 1 & 71.9/63.9 & 65.8/61.3 & 76.0/75.5 & 68.3/67.9 & 71.2/58.7 & 63.1/51.7 & 68.6/66.9 & 66.1/64.4 \\
            CenterPoint~\cite{yin2021center} & CVPR'21 & 1 & -/- & -/- & 76.7/76.2 & 68.8/68.3 & 79.0/72.9 & 71.0/65.3 & -/- & -/- \\
            % RangeDet~\cite{fan2021rangedet} & ICCV'21 & 1 & 71.5/69.5 & 65.0/63.2 & 72.9/72.3 & 64.0/63.6 & 75.9/71.9 & 67.6/63.9 & 65.7/64.4 & 63.3/62.1 \\
            SST~\cite{fan2022embracing} & CVPR'22 & 1 & 74.5/71.0 & 67.8/64.6 & 74.2/73.8 & 65.5/65.1 & 78.7/69.6 & 70.0/61.7 & 70.7/69.6 & 68.0/66.9 \\
            VoxSet~\cite{he2022voxel} & CVPR'22 & 1 & 75.4/72.2 & 69.1/66.2 & 74.5/74.0 & 66.0/65.6 & 80.0/72.4 & 72.5/65.4 & 71.6/70.3 & 69.0/67.7 \\
            % SST-TS~\cite{fan2022embracing} & CVPR'22 & 1 & -/- & -/- & 76.2/75.8 & 68.0/67.6 & 81.4/74.0 & 72.8/65.9 & -/- & -/- \\
            SWFormer~\cite{sun2022swformer} & ECCV'22 & 1 & -/- & -/- & 77.8/77.3 & 69.2/68.8 & 80.9/72.7 & 72.5/64.9 & -/- & -/- \\
            PillarNet-34~\cite{shi2022pillarnet} & ECCV'22 & 1 & 77.3/74.6 & 71.0/68.5 & 79.1/78.6 & 70.9/70.5 & 80.6/74.0 & 72.3/66.2 & 72.3/71.2 & 69.7/68.7 \\
            CenterFormer~\cite{zhou2022centerformer} & ECCV'22 & 1 & 75.3/72.9 & 71.1/68.9 & 75.0/74.4 & 69.9/69.4 & 78.6/73.0 & 73.6/68.3 & 72.3/71.3 & 69.8/68.8 \\
            PV-RCNN++~\cite{shi2023pv} & IJCV'22 & 1 & 78.1/75.9 & 71.7/69.5 & 79.3/78.8 & 70.6/70.2 & 81.3/76.3 & 73.2/68.0 & 73.7/72.7 & 71.2/70.2 \\
            Voxel-DETR~\cite{zhu2022conquer} & CVPR'23 & 1 & 74.9/72.0 & 68.8/66.1 & 75.4/74.9 & 67.8/67.2 & 77.6/70.5 & 69.7/63.1 & 71.7/70.5 & 69.0/67.9 \\
            \midrule
            3D-MAN~\cite{yang20213d} & CVPR'21 & 16 & -/- & -/- & 74.5/74.0 & 67.6/67.1 & 71.7/67.7 & 62.6/59.0 & / & -/- \\
            CenterPoint~\cite{yin2021center} & CVPR'21 & 4 & 76.4/74.9 & 70.8/69.4 & 76.7/76.2 & 69.1/68.6 & 78.9/75.6 & 71.7/68.6 & 73.7/73.0 & 71.6/70.9 \\
            PillarNet-34~\cite{shi2022pillarnet} & ECCV'22 & 2 & 77.6/76.2 & 71.8/70.4 & 80.0/79.5 & 72.0/71.5 & 82.5/79.3 & 75.0/72.0 & 70.5/69.7 & 68.4/67.6 \\
            CenterFormer~\cite{zhou2022centerformer} & ECCV'22 & 2 & 78.3/76.7 & 74.3/72.8 & 77.0/76.5 & 72.1/71.6 & 81.4/78.0 & 76.7/73.4 & 76.6/75.7 & 74.2/73.3 \\
            CenterFormer~\cite{zhou2022centerformer} & ECCV'22 & 4 & 78.5/77.0 & 74.7/73.2 & 78.1/77.6 & 73.4/72.9 & 81.7/78.6 & 77.2/74.2 & 75.6/74.8 & 73.4/72.6 \\
            CenterFormer~\cite{zhou2022centerformer} & ECCV'22 & 8 & 78.7/77.3 & 75.1/73.7 & 78.8/78.3 & 74.3/73.8 & 82.1/79.3 & 77.8/75.0 & 75.2/74.4 & 73.2/72.3 \\
            MPPNet${}^\dagger$~\cite{chen2022mppnet} & ECCV'22 & 4 & 81.1/79.9 & 75.4/74.2 & 81.5/81.1 & 74.1/73.6 & 84.6/82.0 & 77.2/74.7 & 77.2/76.5 & 75.0/74.4 \\
            MSF${}^\dagger$~\cite{He_2023_CVPR} & CVPR'23 & 4 & 81.6/80.2 & 75.9/74.6 & 81.3/80.8 & 73.8/73.3 & 85.0/82.1 & 77.9/75.1 & 78.4/77.6 & 76.1/75.4 \\
            MSF${}^\dagger$~\cite{He_2023_CVPR} & CVPR'23 & 8 & 82.2/80.7 & 76.8/75.5  & 82.8/82.0 & 75.8/75.3 & 85.2/82.2 & 78.3/75.6 & 78.5/77.7 & 76.3/75.5  \\
            \midrule
            STEMD & - & 4 & 81.6/80.0 & 76.1/74.5 & 79.8/79.3 & 72.4/72.0 & 84.5/81.2 & 78.0/74.7 & 80.4/79.5 & 78.0/76.9 \\
            STEMD & - & 8 & 81.9/80.4 & 76.6/75.0 & 80.3/79.8 & 73.2/72.7 & 84.7/81.4 & 78.5/75.2 & \textbf{80.7}/\textbf{79.8} & \textbf{78.3}/\textbf{77.2} \\
            STEMD+MSF${}^\dagger$ & - & 8  & \textbf{83.1}/\textbf{81.5}   & \textbf{77.7}/\textbf{76.2}  & \textbf{83.2}/\textbf{82.8} & \textbf{75.8}/\textbf{75.6} & \textbf{85.8}/\textbf{82.5} & \textbf{79.0}/\textbf{75.9} & 80.2/79.2 & 78.2/77.1   \\
            \bottomrule
		\end{tabular}
	}
\end{table*}

\setlength{\tabcolsep}{8pt}
\begin{table}
	\centering
	%	The results of 3D object detection on Waymo Open Dataset validation set (train with 100% training data).
	\caption{\revise{Comparison with state-of-the-art methods on the validation set of nuScenes Dataset.}}
	\label{table:nuscenes_val}
	\begin{tabular}{c|c|c|cc}
	\toprule
	Method & Publication & Input (s) & mAP & NDS \\
	\midrule
	TransPillars~\cite{luo2023transpillars} & WACV'23 & 1.5 & 52.3 & - \\
	CenterPoint~\cite{yin2021center}        & CVPR'21 & 0.5 & 55.6 & 63.5 \\
	Focals Conv~\cite{chen2022focal}        & CVPR'22 & 0.5 & 61.2 & 68.1 \\
	TransFusion-L~\cite{bai2022transfusion} & CVPR'22 & 0.5 & 65.1 & 70.1 \\
	\midrule
	Ours (STEMD)                            & -       & 1.5 & 67.5 & 71.6 \\
	\bottomrule
	\end{tabular}
\end{table}
\setlength{\tabcolsep}{1pt}

\subsection{Implementation Details}\label{sec:implementation_details}
\noindent\textbf{Network Architectures.} 
\revise{
First, we employ average pooling to derive voxel-wise feature maps by encoding the point cloud within each voxel. Our method incorporates a ResNet18~\cite{he2016deep} 3D backbone, modifying the 2D convolution modules to their 3D counterparts. 
%For computational efficiency, submanifold sparse convolution layers are used in all residual blocks, except in the down-sampling layers. 
To augment bird's eye view (BEV) features, an FPN structure~\cite{lin2017feature} is implemented. The 8$\times$ down-sampled BEV feature maps are processed through a 3-layer encoder, integrating BoxAttention~\cite{nguyen2022boxer}, a type of Deformable Attention~\cite{zhu2021deformable}, for self-attention mechanisms. 
In the following ConvGRU module, a learnable $3\times 3$ kernel is utilized in all convolution layers. This module's output retains the same channel configuration as the input BEV features. Enhanced BEV features from the ConvGRU are exploited using a class-agnostic FFN head for object proposal generation~\cite{zhu2022conquer}, and we employ a CenterHead~\cite{yin2021center} specifically for the nuScenes dataset. The decoder employs standard multi-head attention for self-attention and BoxAttention for cross-attention between queries and BEV maps. Both the encoder and decoder are set with a hidden size of 256 and eight attention heads. 
During training, the temporal query recollection strategy is adopted. It involves selecting the top 300 highest-scoring predicted results from the previous frame to initialize the object queries in the decoder, supplementing the top 1000 scored proposals from the current frame's encoder. In graph node selection, we set the number of bounding boxes, $N_g$, to 800 and the IoU threshold, $\theta$, to 0.5. The spatial-temporal graph attention network employs distance thresholds $d_s = 2\,\text{m}$ and $d_u = 2\,\text{m}$ for the spatial self-attention module and temporal graph-to-graph attention network, respectively. 
For the loss function, we assign the weights $\lambda_{\text{cls}} = 1$, $\lambda_{\text{h}} = 4$, $\lambda_{\text{giou}} = 2$, and $\lambda_{\text{r}} = 1$ for the decoder's prediction head and graph head, while setting it to 0 for the encoder's proposal head. The results of our method in Table~\ref{table:waymo_val} use 2x wider ResNet and contrastive denoising training~\cite{zhang2023dino}. For the combination of STEMD and MSF, we also estimate the speeds of the detected objects in the first phase.
}\\

\noindent\textbf{Training Details.}
\revise{
For experiments on the Waymo dataset, input points are selected within coordinates \([-75.2\,\text{m}, 75.2\,\text{m}]\) on both the \(x\) and \(y\) axes, and \([-2\,\text{m}, 4\,\text{m}]\) along the \(z\) axis. Point clouds are voxelized with a grid size of \((0.1\,\text{m}, 0.1\,\text{m}, 0.15\,\text{m})\). During training and inference, the maximum number of non-empty voxels is capped at 15000. The Adam optimizer is employed, with \(\beta_1 = 0.9\) and \(\beta_2 = 0.99\), and an initial learning rate of 0.003, which is updated using a one-cycle scheduler~\cite{smith2017cyclical} and a weight decay of 0.01. The model undergoes a total of 12 epochs of end-to-end training. A score threshold of 0.1 is set to filter out low-quality predictions during inference. For the nuScenes dataset, the input range on the $x$-$y$ plane is \([-54.0\,\text{m}, 54.0\,\text{m}]\), and \([-5\,\text{m}, 3\,\text{m}]\) on the $z$-axis. Ego-motion correction is applied to account for the self-motion of the vehicle. Points from associated sweeps are concatenated with the keyframe to form a single input frame. Following the methodologies in ~\cite{luo2023transpillars, yin2021graph}, 3 keyframes and their associated sweeps are used as input. The detector is trained over 20 epochs with an initial learning rate of 0.001, updated using a one-cycle scheduler.
}

\subsection{Comparison with State-of-the-Art Methods}\label{sec:waymo_val_result}
\setlength{\tabcolsep}{0.5pt}

We compare our STEMD with the current state-of-the-art methods for 3D object detection. It can be observed in Table~\ref{table:waymo_val} that approaches employing multiple frames generally outperform those using a single frame as input.
When comparing our method to the previous state-of-the-art single-frame method, PV-RCNN++, our STEMD achieves a significant improvement in the overall 3D mAPH by utilizing an 8-frame point cloud sequence as input. Specifically, we observe remarkable enhancements in the detection performance for vehicles, pedestrians, and cyclists, with respective improvements of 2.5\% APH, 7.2\% APH, and 7.0\% APH on LEVEL\_2.
Comparing our approach to previous single-frame detectors, the progress achieved by STEMD validates its capability to successfully leverage the spatiotemporal dependencies, aiding in the accurate estimation of objects that are challenging to detect in the setting of single-frame input.
Moreover, our method also demonstrates superiority when compared to the previous multi-frame approaches like CenterFormer~\cite{zhou2022centerformer}. By employing our method, we are able to achieve a notable improvement of 4.9\% APH on LEVEL\_2 for the cyclist category.
Additionally, our method exhibits improved performance when using shorter input sequences. Notably, even with a 4-frame input, our approach outperforms both the current single-frame and multi-frame methods. Specifically, STEM-4f shows superior performance compared to PV-RCNN++ and MPPNet-4f, with improvements of 5.5\% and 0.8\% mAPH on LEVEL\_2, respectively.
Furthermore, it is important to highlight that the single STEMD method is an \textit{end-to-end} solution and does not rely on speed supervision, distinguishing it from certain high-accuracy but two-phase methods~\cite{chen2022mppnet,He_2023_CVPR,qi2021offboard}. 
\major{The MSF and MPPNet are plug-and-play to existing 3D detectors, because they can further refine the preliminary 3D bounding boxes by associating the 3D proposals across frames. Thus, we further integrate STEMD with MSF and achieve the best results with an mAPH of 81.5\% on LEVEL\_1 and 76.2\% on LEVEL\_2. This combination surpasses the previous state-of-the-art achieved by MSF alone, which recorded 80.7\%  mAPH on LEVEL\_1 and 75.5\% mAPH on LEVEL\_2.}
We also evaluate our method on the nuScenes dataset, where our STEMD outperforms the current start-of-the-art single/multi-frame 3D detection method by a margin of 2.4\% mAP and 1.5\% NDS.
Overall, the results in Table~\ref{table:waymo_val} and Table~\ref{table:nuscenes_val} highlight the efficacy of STEMD in effectively integrating temporal information from point cloud sequences, leading to enhanced detection accuracy compared to existing single-frame and multi-frame methods.

\subsection{Ablation Studies}\label{sec:ablation_study}
We conduct ablative analyses to verify the effectiveness and characteristics of our processing pipeline. In the absence of special instructions, we use 4-frame point cloud input unless otherwise specified in this part. 
We report the APH on LEVEL\_2 for vehicle, pedestrian, and cyclist categories and the mAPH metric on the Waymo dataset for comparison.

\setlength{\tabcolsep}{6pt}
\begin{table}[t]
	\centering
	\caption{Contribution of each component in the proposed method. The results of mAPH on LEVEL\_2 are reported. ``S.I." denotes the sequential point cloud input. ``S.T." denotes spatial-temporal graph attention network. ``TQR" means the temporal query recollection strategy. ``IoU Reg" is the IoU regularization term in the loss function.}
	\label{table:ablation_components}
	\begin{tabular}{cccc|ccc>{\columncolor{whitesmoke}}c} 
		\specialrule{1pt}{0pt}{0pt}
		S.I. & S.T. & TQR & IoU Reg. & Veh.  & Ped.  & Cyc. & Mean   \\ 
		\hline
			$\times$    & $\times$  & $\times$ & $\times$   & 68.33 & 67.36 & 71.16 & 69.62  \\
			\checkmark     & $\times$ & $\times$ & $\times$ & 68.41 & 69.82 & 72.06 & 70.10  \\
			\checkmark     & \checkmark  & $\times$ & $\times$ & 68.71 & 72.10 & 75.64 & 72.15  \\
			\checkmark     & \checkmark  & \checkmark  & $\times$ & 69.30 & 73.30 & 76.01 & 72.87  \\
			\checkmark     & \checkmark  & \checkmark  & \checkmark  & 70.83 & 73.62 & 76.80 & 73.75  \\
		\specialrule{1pt}{0pt}{0pt}
	\end{tabular}
\end{table}
\setlength{\tabcolsep}{5pt}

\noindent\textbf{Effect of Key Components.}
In Table~\ref{table:ablation_components}, we investigate the effect of each added component in our method on the setting of 4-frame sequence input.
% Sequential processing, graph selection, spatial-temporal GAN, Temporal Query Recollection, IoU regularization term.
As can be seen from the $\text{1}^{\text{st}}$ row and $\text{2}^{\text{nd}}$ row in Table~\ref{table:ablation_components}, after changing the processing of multi-frame point cloud input from direct concatenation to operating it as a sequence, the results suggest an improvement and reaches 70.1\% LEVEL\_2 mAPH. This indicates that compared with the simple concatenation, the temporal feature-level enhancement by the ConvGRU block benefits the network in capturing the spatial-temporal dependencies. This also motivates us to take full advantage of the point cloud sequence and capture the spatial-temporal dependencies at not only the feature level but also the query level.
When the spatial-temporal graph attention network is applied, we can further improve the result of vehicle, pedestrian, and cyclist by 0.3\%, 2.28\%, and 3.58\% LEVEL\_2 mAPH respectively. We attribute this improvement to the interaction patterns exploited by the graph attention network, which effectively captures valuable information regarding the social behaviors exhibited by objects. This mechanism proves influential in complex scenarios, displaying a pronounced impact on categories such as pedestrians and cyclists. As reported in the fourth row of Table~\ref{table:ablation_components}, the strategy of temporal query recollection yields an improvement of 0.72\% mAPH, indicating the effectiveness of supplementing the queries initialized with boxes predicted by the encoder of the current frame with the final prediction in last frame. As shown in the last row of Table~\ref{table:ablation_components}, by introducing the IoU regularization term in the loss function of FFN prediction head, the performance will increase by 0.88\% LEVEL\_2 mAPH.\\

\setlength{\tabcolsep}{8pt}
\begin{table}
	\centering
	\caption{
        The results comparison of our STEMD with the baseline~\cite{zhu2022conquer} using point concatenation strategy under different frame length settings. We report the results of APH on LEVEL\_2 for three categories.}
	\label{table:ablation_sequential_processing}
	\begin{tabular}{c|c|ccc>{\columncolor{whitesmoke}}c} 
		\specialrule{1pt}{0pt}{0pt}
		Frames              & Method      & Veh.  & Ped.  & Cyc.  & Mean   \\ 
		\hline
		\multirow{3}{*}{2}  & Concat.     & 68.12 & 67.74 & 71.38 & 69.08  \\
		& Ours     & 69.03 & 69.82  & 72.45 & 70.43 \\
		& \textit{Improvement} & +0.91  & +2.08  & +1.07  & +1.35   \\ 
		\hline
		\multirow{3}{*}{4}  & Concat.  &   68.33 &	69.36 &	71.16 &	69.62 \\
		& Ours        & 70.83 & 73.62 & 76.80 & 73.75  \\
		& \textit{Improvement} & +2.50	  & +4.26	  & +5.64	  & +4.13 \\ 
		\hline
		\multirow{3}{*}{8}  & Concat.  &  69.25 &	69.76 &	70.02 &	69.68 \\
		& Ours        & 71.74 & 74.09 & 77.07 & 74.30  \\
		& \textit{Improvement} &	+2.49	& +4.33	& +7.05	& +4.62  \\ 
		\hline
		\multirow{3}{*}{16} & Concat.     & 69.12 &	69.82 &	70.02 &	69.65  \\
		& Ours        & 72.10  & 74.37 & 77.22 & 74.56  \\
		& \textit{Improvement} & +2.98 &	+4.55 &	+7.20 &	+4.91 \\
		\specialrule{1pt}{0pt}{0pt}
	\end{tabular}
\end{table}
\setlength{\tabcolsep}{8pt}

\noindent\textbf{Effect of our Multi-frame Design.}
As shown in Table~\ref{table:ablation_sequential_processing}, we show the improvement of our multi-frame design compared to the baseline~\cite{zhu2022conquer} using simple point cloud concatenation strategy under different settings of frame length.
We can observe that the baseline does not perform well in modeling the long-term relations among frames as the performance slightly deteriorates when the number of concatenated frames increases from 8 to 16. In contrast, our proposed STEMD achieves consistent improvements when the frame length increases. Specifically, our proposed STEMD achieves 1.35\%, 4.13\%, 4.62\%, and 4.91\% higher mAPH than the multi-frame baseline using point concatenation with 2, 4, 8, and 16 frames. The results demonstrate the effectiveness of our method in leveraging long-term temporal dependencies across multiple frames.\\

\setlength{\tabcolsep}{8pt}
\begin{table}[t]
	\centering
	\caption{Comparison of STGA-Net and conventional multi-head self-attention. We report the results of APH on LEVEL\_2 for three categories.}
	\label{table:ablation_attention}
	\begin{tabular}{l|ccc>{\columncolor{whitesmoke}}c} 
		\specialrule{1pt}{0pt}{0pt}
		Setting        & Veh.  & Ped.  & Cyc.  & Mean   \\ 
		\hline
		STGA-Net         & 70.83 & 73.62 & 76.80 & 73.75  \\
		Self-Attention & 70.60 & 71.24 & 73.71 & 71.85  \\ 
		\hline
		Diff           & -0.23 & -2.38 & -3.09 & -1.90  \\
		\specialrule{1pt}{0pt}{0pt}
	\end{tabular}
\end{table}
\setlength{\tabcolsep}{8pt}

\setlength{\tabcolsep}{8pt}
\begin{table}[t]
	\centering
	\caption{
		\major{Effect of the \textit{spatial} self-attention and \textit{temporal} cross-attention module in the proposed STGA-Net. The results of mAP and mAPH on LEVEL\_1 and LEVEL\_2 are reported.}
	}
	\label{table:ablation_spatial_temporal}
	\begin{tabular}{cc|cccc}
		\specialrule{1pt}{0pt}{0pt}
		\multirow{2}{*}{Spatial~} & \multirow{2}{*}{Temporal} & \multicolumn{2}{c}{LEVEL\_1} & \multicolumn{2}{c}{LEVEL\_2}  \\
			 &  	& mAP  	& mAPH  & mAP   & mAPH \\
			 \hline
		$\times$    & $\times$    & 78.47 & 77.12    & 72.98 & 71.56     \\
		$\times$    & \checkmark    & 78.69 & 77.25    & 73.11 & 71.69     \\
		\checkmark    & $\times$    & 79.27 & 77.82    & 73.79 & 72.34     \\
		\checkmark    & \checkmark    & 80.58 & 79.09    & 75.22 & 73.75    \\
		\specialrule{1pt}{0pt}{0pt}
	\end{tabular}
\end{table}
\setlength{\tabcolsep}{8pt}

\noindent\textbf{Effect of the Spatial-temporal Graph Attention Network.}
To showcase the effectiveness of the spatial-temporal graph attention network, we conduct additional experiments where we replace it with conventional multi-head self-attention modules~\cite{vaswani2017attention}. The results, as presented in Table~\ref{table:ablation_attention}, demonstrate that the normal multi-head self-attention mechanism performs 0.23\%, 2.38\%, and 3.09\% worse in LEVEL\_2 mAPH for cars, pedestrians, and bicycles categories, respectively, compared to the proposed spatial-temporal graph attention network. This indicates that the complex spatial-temporal dependencies between objects are more effectively modeled by the graph structure, which dynamically and contextually captures the relationships, as opposed to the self-attention mechanism that is densely applied across all queries. \major{To evaluate the impact of the individual components within the STGA-Net, we conducted experiments to isolate the effects of the spatial self-attention module and the temporal cross-attention module (see Table~\ref{table:ablation_spatial_temporal}). Enabling only the spatial self-attention or temporal cross-attention module showed a slight improvement. The highest performance was observed when both modules were enabled, with mAPH scores of 79.09\% (L1) and 73.75\% (L2), demonstrating the combined effectiveness of modeling both spatial and temporal dependencies for multi-frame 3D object detection.\\}

\begin{figure}[t]
	\centering
	\includegraphics[width=0.49\textwidth]{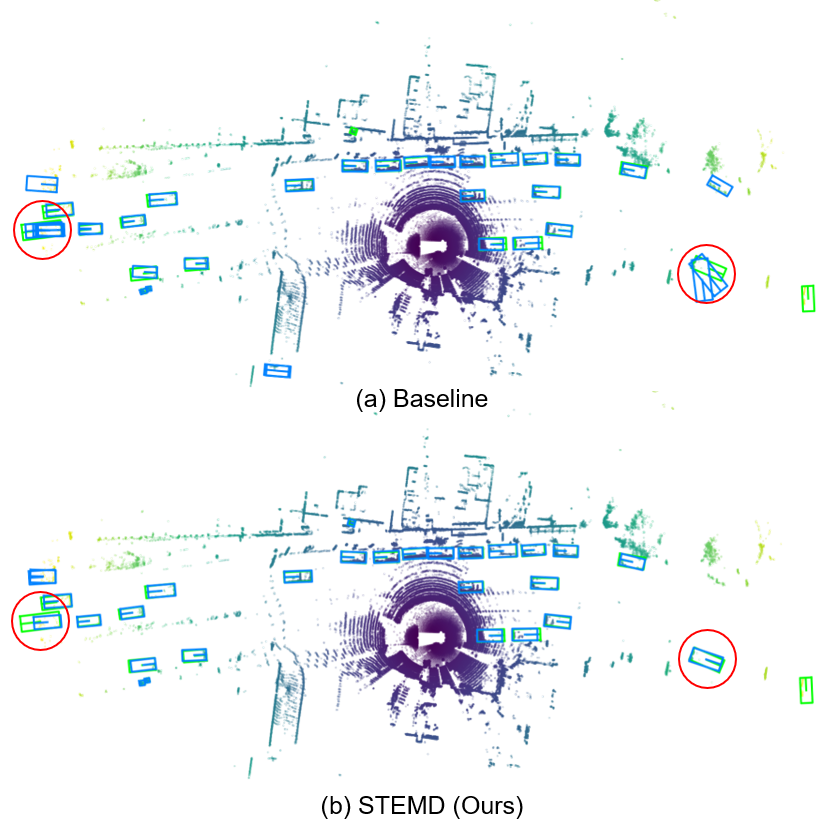} 
	\caption{
		Visual comparison of the full model of STEMD (right) with the baseline setting w/o IoU regularization term. The green boxes represent the ground truth bounding boxes, while the blue boxes indicate the bounding box predictions with confidence larger than 0.1. The results of the baseline contain several overlapped boxes, while the full model generates sparser predictions (highlighted in the red circles).
	} 
	\label{fig:vis_iou_reg}
\end{figure}

\noindent\textbf{Effect of the IoU Regularization Term.}
As shown in Fig.~\ref{fig:vis_iou_reg}, we conduct a qualitative comparison between the results of the full model of STEMD (shown on the right side) and the baseline setting without the IoU regularization term (left side). Upon analysis, we observe that the predictions of the baseline model exhibited highly overlapped bounding boxes, which are highlighted with red circles. In contrast, the full model produced a single matched box for each object. 
This observation suggests that the introduction of the regularization term in the full model helps to push other close but unmatched queries away from the best-matched queries during the refinement in the decoder, and consequently leads to less-overlapped box predictions.\\

\noindent\textbf{Hyper-parameters of Graph Node Selection.}
\setlength{\tabcolsep}{5pt}
\begin{table}
	\centering
	\caption{Effect of the number of nodes $N_{g}$ adopted in STGA-Net. We report the recall rate of nodes in the graph with respect to the ground-truth bounding boxes at IoU thresholds of 0.7 and 0.5. We also report the results of APH on LEVEL\_2 for three categories.}
	\label{table:ablation_graph_node_number}
	\begin{tabular}{c|cc|ccc>{\columncolor{whitesmoke}}c} 
		\specialrule{1pt}{0pt}{0pt}
		$N_{g}$ & Recall@0.7 & Recall@0.5 & Veh.  & Ped.  & Cyc.  & Mean   \\ 
		\hline
		1300     & \textbf{88.11}      & \textbf{97.05}      & 70.47 & 72.23 & 75.22 & 72.64  \\
		1000     & 87.68      & 97.04      & \textbf{70.86} & 73.35 & 76.73 & 73.65  \\
		800      & 87.10      & 96.95      & 70.83 & 73.62 & \textbf{76.80} & \textbf{73.75}  \\
		500      & 85.75      & 96.56      & 70.03 & 73.65 & 76.75 & 73.47  \\
		300      & 84.64      & 96.10      & 69.87 & \textbf{73.69} & 75.39 & 72.98  \\
		\specialrule{1pt}{0pt}{0pt}
	\end{tabular}
\end{table}
\setlength{\tabcolsep}{2pt}
In our approach, we incorporate graph node selection to address the issue of overlapping bounding boxes and associated queries generated by the decoder. This process results in a more streamlined graph structure for downstream graph-based learning. However, it is crucial to determine the appropriate number of filtered bounding boxes, denoted as $N_{g}$, as an excessively large value can introduce redundant bounding boxes that impede the effectiveness of graph-based learning. Conversely, setting $N_{g}$ too small may result in a lower recall rate of queries with respect to the ground-truth boxes. By analyzing Table~\ref{table:ablation_graph_node_number}, we observe a significant decrease in the recall rate at the 0.5 IoU threshold, dropping from 96.95\% to 96.56\% when $N_{g}$ is reduced from 800 to 500. Thus, we find that setting $N_{g}$ to 800 strikes a balance between recall and redundancy within the graph, thereby leading to the best detection accuracy of 73.75\% mAPH.\\

\noindent\textbf{Hyper-parameters of Temporal Query Recollection.}
\setlength{\tabcolsep}{6pt}
\begin{table}[t]
	\centering
	\caption{Effect of the number of queries in decoder initialized by the box predictions in the last frame, i.e., $N_{\text{res}}$. We report the recall rate of all queries input for the decoder with respect to the ground-truth bounding boxes at IoU thresholds of 0.7 and 0.5. We also report the results of APH on LEVEL\_2 for three categories.}
	\label{table:ablation_recollection}
	\begin{tabular}{c|cc|ccc>{\columncolor{whitesmoke}}c} 
		\specialrule{1pt}{0pt}{0pt}
		$N_{\text{res}}$ & Recall@0.7  & Recall@0.5   & Veh. & Ped. & Cyc. & Mean  \\ 
		\hline
		0          & 68.89    & 88.80    & 70.28 & 72.66 & 76.41 & 73.12  \\
		100        & 69.06    & 89.00    & 70.51 & 72.95 & 76.49 & 73.32  \\
		200        & 69.45    & 89.22    & 70.74 & 73.50 & 76.55 & 73.60  \\
		300        & 69.75    & 89.31    & \textbf{70.83} & 73.62 & 76.80 & \textbf{73.75}  \\
		500        & 69.76    & 89.36    & 70.79 & 73.64 & 76.82 & \textbf{73.75}  \\
		800        & 69.81    & 89.39    & 70.71 & \textbf{73.66} & \textbf{76.85} & 73.74  \\
		\specialrule{1pt}{0pt}{0pt}
	\end{tabular}
\end{table}
\setlength{\tabcolsep}{5pt}
As shown in Table~\ref{table:ablation_recollection}, with extra $N_{\text{res}}$ queries initialized with boxes predicted at the last frame, both the recall of all queries input for decoder with respect to the ground-truth bounding boxes and the overall mAPH improves. Specifically, the recall increases from 68.89\% to 69.75\% and from 88.80\% to 89.31\% at 0.7 and 0.5 IoU thresholds, respectively. But when we further increase the $N_{\text{res}}$ from 300 to 500 or 800, neither the recall nor the detection accuracy improves significantly. Therefore, we can infer the collected queries in the last frame could effectively supplement the output of encoder in current timestamp as extra input queries of decoder. And the value of $N_{\text{res}}$ is important because too many recollected queries are helpless and could lead to unnecessary computation load.\\

\setlength{\tabcolsep}{8pt}
\begin{table}[htbp]
	\centering
	\caption{
		\major{Comparison of different edge weights obtained based on the Euclidean distance and learned edge weights based on node features. The results of mAP and mAPH on LEVEL\_1 and LEVEL\_2 are reported.}
	}
	\label{table:ablation_edge_weights}
	\begin{tabular}{c|cccc}
		\specialrule{1pt}{0pt}{0pt}
		\multirow{2}{*}{Edge Weights} & \multicolumn{2}{c}{LEVEL\_1} & \multicolumn{2}{c}{LEVEL\_2}  \\
		  	& mAP  	& mAPH  & mAP   & mAPH \\
		\hline
		 Distance-based    		   & 79.19 & 77.54    & 73.75 & 71.17    \\
		 Feature-based			   & 80.58 & 79.09    & 75.22 & 73.75    \\
		\specialrule{1pt}{0pt}{0pt}
	\end{tabular}
\end{table}
\setlength{\tabcolsep}{8pt}

\noindent\textbf{\major{Effect of Different Edge Weights Calculations.}} 
\major{Table~\ref{table:ablation_edge_weights} presents the comparison of different methods for calculating edge weights in the spatial-temporal graph attention network. We evaluated two approaches: distance-based edge weights calculated using the Euclidean distance between nodes, and feature-based edge weights learned from node features. The results show that using distance-based edge weights yields mAPH scores of 77.54\% on LEVEL\_1 and 71.17\% on LEVEL\_2. In contrast, feature-based edge weights significantly improve performance, achieving mAPH scores of 79.09\% on LEVEL\_1 and 73.75\% on LEVEL\_2. These findings indicate that learning edge weights based on node features provides a more effective and adaptive representation of spatial-temporal dependencies, leading to better detection accuracy compared to using static distance-based calculations.\\}

\setlength{\tabcolsep}{6pt}
\begin{table}[htbp]
	\centering
	\caption{
		\major{Effect of varying levels of graph sparsity by adjusting the largest number of connections each node maintains. The results of mAP and mAPH on LEVEL\_1 and LEVEL\_2 are reported.}
	}
	\label{table:ablation_edge_number}
	\begin{tabular}{c|cccc}
		\specialrule{1pt}{0pt}{0pt}
		\multirow{2}{*}{Largest Edge Number} & \multicolumn{2}{c}{LEVEL\_1} & \multicolumn{2}{c}{LEVEL\_2}  \\
		& mAP  	& mAPH  & mAP   & mAPH \\
		\hline
		5  &	78.53 &	77.06 &	73.19 &	71.74    \\
		10 & 79.40 &	77.86 &	73.98 &	72.46    \\
		15 & 79.87 &	78.38 &	74.56 &	73.09	\\
		20 & 80.16 &	78.57 &	75.01 &	73.42	\\
		No Limit  &	80.58 &	79.09 &	75.22 &	73.75	\\
		\specialrule{1pt}{0pt}{0pt}
	\end{tabular}
\end{table}
\setlength{\tabcolsep}{8pt}

\noindent\textbf{\major{Effect of the Graph Sparsity.}} \major{Table~\ref{table:ablation_edge_number} shows the impact of varying graph sparsity by adjusting the maximum number of connections each node maintains. The performance deteriorates significantly when we limit the number of edges per node to 15 and lower. With 5 edges per node, the model achieves mAPH scores of 77.06 and 71.74 on LEVEL\_1 and LEVEL\_2, respectively. These results indicate that allowing connections of adapted number per node better captures complex spatial-temporal relationships, while overly limiting the connections significantly reduces performance.}\\

\begin{figure}[t]
	\centering
	\includegraphics[width=0.48\textwidth]{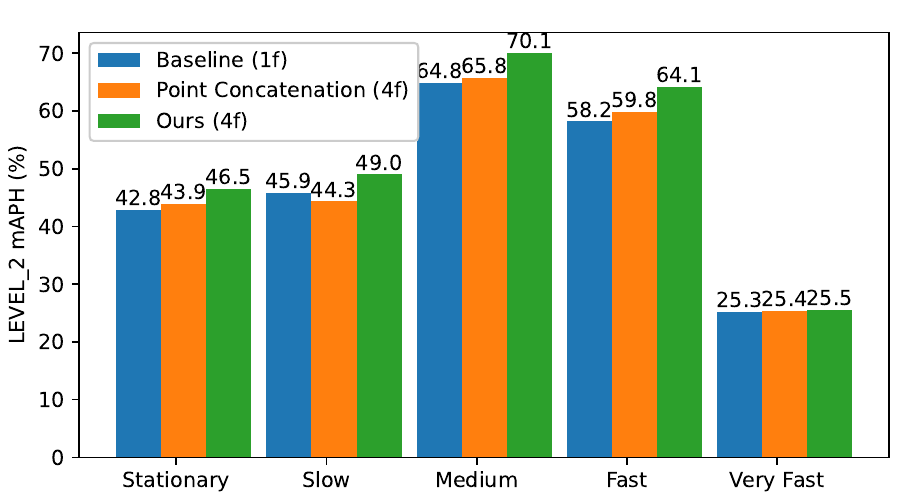} 
	\caption{
		The performance (mAPH) breakdown over different speeds. As defined by the official evaluation tools
		of Waymo dataset, objects can be classified based on their speed as follows: stationary (less than 0.2 m/s), slow (0.2 to 1 m/s), medium (1 to 3 m/s), fast (3 to 10 m/s), or very fast (greater than 10 m/s).
	} 
	\label{fig:speed_breakdown}
\end{figure}
%\vspace{-1cm}

\begin{figure}[htbp]
	\centering
	\includegraphics[width=0.49\textwidth]{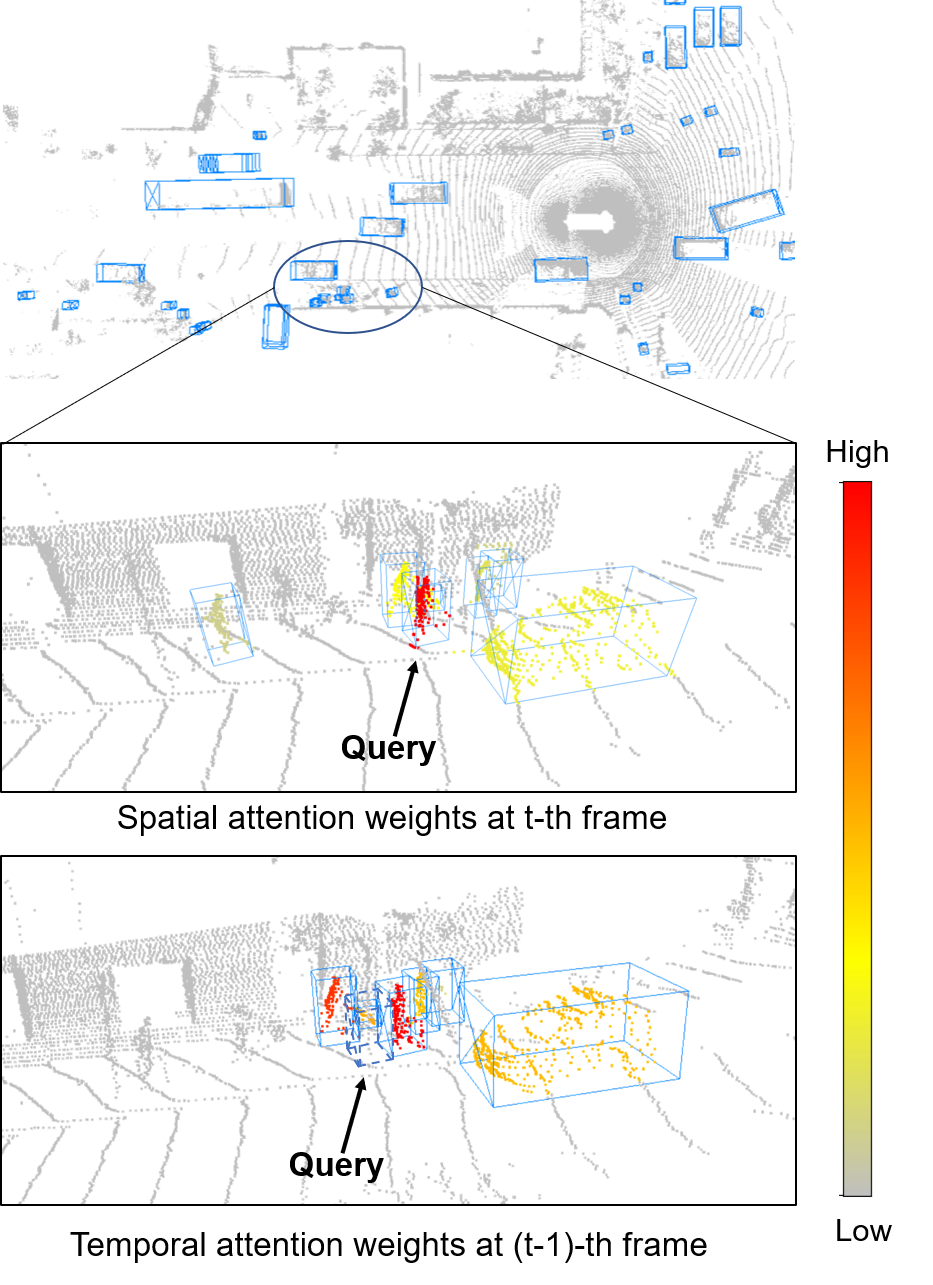} 
	\caption{
		\revise{
			Illustration of attention weights in the spatial-temporal graph attention network. This figure illustrates the spatial attention weights for associated objects in relation to a target query within the spatial graph at the current frame, and also shows the attention coefficients relative to the graph node from the last frame.
		}
	}
	\label{fig:attention_map}
\end{figure}

\begin{figure*}[htbp]
	\centering
	\includegraphics[width=0.95\textwidth]{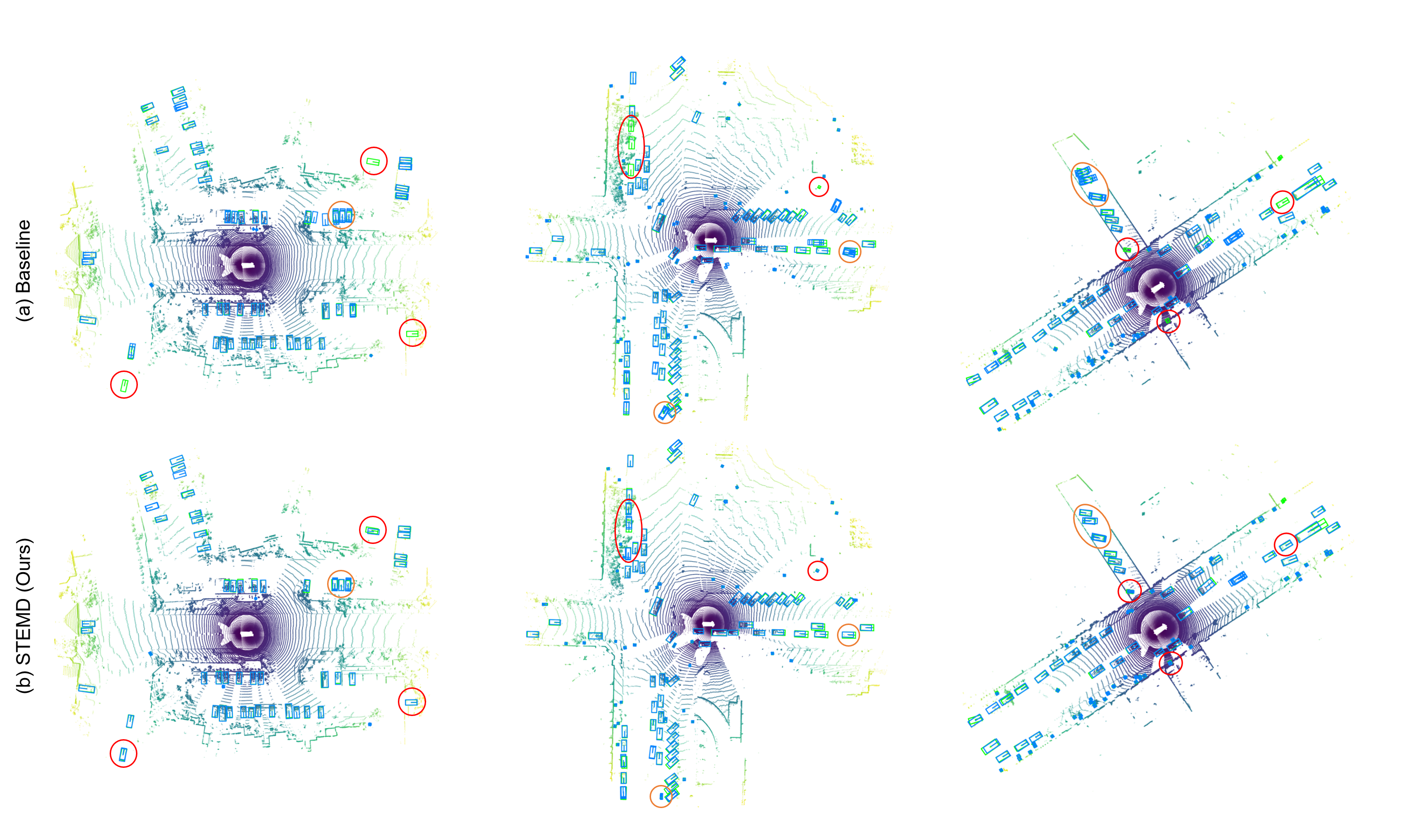} 
	\caption{
		\revise{Visual comparison between the results obtained by the multi-frame baseline~\cite{zhu2022conquer} method utilizing point concatenation and our proposed STEMD on the Waymo Open Dataset validation set. The ground-truth bounding boxes are shown in \textcolor{green}{green}, while the predicted bounding boxes are depicted in \textcolor{blue}{blue}. We employ \textcolor{red}{red} circles to highlight objects that are successfully detected by STEMD but missed by the baseline approach. Additionally, we use \textcolor{orange}{orange} circles to emphasize certain objects where the baseline method produces false-positive or redundant bounding boxes, while STEMD predicts more precise bounding boxes. Best viewed in color and zoom in for more details.}
	}
	\label{fig:vis_waymo_val}
\end{figure*}

\noindent\textbf{Conditional Analysis.}
To better figure out where our approach brings improvements, we compare STEMD with the baseline that simply concatenates the multi-frame point clouds on objects of different speeds in Fig.~\ref{fig:speed_breakdown}. 
We can observe that the baseline model, based on the concatenation of 4 frames, achieves marginal improvement compared with the single-frame detector, and the performance on the slow category even deteriorates. Meanwhile, our proposed STEMD achieves consistent improvements throughout all categories. Especially, our method outperforms the single-frame baseline by 5.3\% and 5.9\% APH improvement on medium and fast objects, the boost is more significant than that on stationary and slow categories. This observation demonstrates our proposed STEMD effectively captures the temporal information about the motion and behavior of fast-moving objects across different frames.

\subsection{Visualization and Analysis}

\revise{We visualize the learned attention patterns in the spatial-temporal graph attention network. As illustrated in Fig.~\ref{fig:attention_map}, we present the attention weights of associated nodes (queries) for a pedestrian on a street. It is observed that the graph attention mechanism effectively focuses on the related queries. The spatial self-attention module models the spatial interactions among objects to avoid collisions and the temporal cross-attention module captures the temporal dependencies like the relevant past location information and the trajectories of an oncoming gang of pedestrians. 
Operation of the traditional self-attention mechanism considers relationships between each query and all other queries, the resulting redundancy of which is circumvented in GATs by focusing on the most relevant interactions.
}

We present a qualitative analysis that highlights the strengths of our STEMD method in multi-frame 3D object detection, compared with the baseline method~\cite{zhu2022conquer} using concatenated points as input, as shown in Fig. \ref{fig:vis_waymo_val}. 
STEMD outperforms the baseline in detecting distant and occluded vehicles, as evident in the scenes of parked cars (marked by red circles in the figure) and reduces false positives and redundant detections (orange circles). 
This demonstrates the effectiveness of STEMD in challenging scenarios, leveraging spatial-temporal dependencies and historical location data.

\setlength{\tabcolsep}{8pt}
\begin{table*}[htbp]
	\centering
	\caption{\major{Comparison of computation cost and latency of detectors on Waymo validation set.}}
	\label{table:latency_comparison}
	%	\scalebox{0.92}{
        \begin{tabular}{l|c|ccccc} 
          \toprule
          Method       & Frames  & Latency (ms) & Memory (MB) & FLOPs (G) & Params (M) & mAP (L2)  \\ 
          \midrule
          CenterPoint~\cite{yin2021center}  & 1 & 39.04        & 2623        & 86.81     & 4.83       & 65.9            \\
          CenterPoint~\cite{yin2021center}  & 4 & 64.57        & 3295        & 88.29     & 4.87       & 70.8            \\
          CenterFormer~\cite{zhou2022centerformer} & 4  & 226.39       & 5349        & 741.60    & 10.76      & 74.7            \\
          MPPNet~\cite{chen2022mppnet}       & 4  & 342.09       & 6877        & 154.51    & 4.78       & 75.4            \\
          MSF~\cite{He_2023_CVPR}          & 4  & 158.90       & 15373       & 83.28     & 3.80       & 75.9            \\ 
          \midrule
          Ours         & 4  & 115.71       & 3861        & 239.46    & 12.93      & 76.1        \\
          \bottomrule    
        \end{tabular}
		%	}
\end{table*}
\setlength{\tabcolsep}{8pt}

\setlength{\tabcolsep}{8pt}
\begin{table*}[htbp]
	\centering
	\caption{\major{The latency breakdown of STEMD on Waymo validation set.}}
	\label{table:latency_decompose}
	%	\scalebox{0.88}{
		\begin{tabular}{c|cccccc|c}
			\toprule
			Latency (\textbf{ms}) & Backbone & Encoder & ConvGRU & Proposal Head & Decoder & STGA-Net & Total   \\
			\midrule
			Baseline-1f  & 42.12    & 25.60   & -       & 2.58          & 6.22    & -        & 76.53   \\
			STEMD-4f     & 42.11    & 25.56   & 36.02   & 2.58          & 6.35    & 3.09     & 115.71 \\
			\bottomrule
		\end{tabular}
		%	}
\end{table*}
\setlength{\tabcolsep}{8pt}

\subsection{Efficiency Analysis}\label{sec:efficiency}

\major{Efficiency is an essential factor for multi-frame 3D object detection algorithms. Therefore, we evaluate the detailed computation cost and runtime of STEMD on the Waymo validation set using a single NVIDIA GeForce RTX A6000 GPU. As shown in Table~\ref{table:latency_comparison}, our proposed STEMD, when processing four frames, demonstrates a latency of 115.71 ms, memory usage of 3861 MB, 239.46 GFLOPs, and 12.93 million parameters, achieving an mAP (L2) score of 76.1\%. We can observe that our method offers the highest mAP (L2) score with lower latency and memory usage compared to most other methods.}

\major{In the online multi-frame 3D object detection using STEMD, frames are processed sequentially in a streaming fashion. To avoid redundant computations, a small set of features computed in the last frame is reused. Therefore, STEMD introduces only a small computational overhead. Table~\ref{table:latency_decompose} provides a detailed latency breakdown of our method. Compared to the single-frame baseline latency of 76.53 ms, our method's additional computational overhead primarily comes from the ConvGRU and STGA-Net modules. Despite this, the increase in latency is acceptable, given the significant improvements in accuracy. Importantly, with a typical LiDAR scan frequency of 10 Hz, our network remains capable of real-time operation within the 100 ms computation budget, with further acceleration in hardware.}

\major{As mentioned in Sec.~\ref{sec:loss}, our method only additionally employs a small set of features derived in the last frame, which means that the additional memory required for storing these features is low. Besides, while some methods directly deal with concatenated multi-frame point clouds, the proposed method only takes the point cloud of the current frame as input. This reduces the number of points that need to be processed for detection, resulting in lower memory requirements during inference. Quantitatively, our method consumes 3861M of GPU memory during inference, which is only slightly higher than the single-frame method's usage of 3470M when the batch size is set to 1. By emphasizing these specific measurements, we can confidently conclude that the STEMD method remains memory efficient.}

\subsection{Further Discussions}
While our experiments have demonstrated the significant benefits of leveraging spatial-temporal information for 3D object detection, it is important to acknowledge the limitations of our current approach. 
One key limitation is that the proposed STEMD falls short in terms of performance compared to some state-of-the-art two-phase methods ~\cite{chen2022mppnet,He_2023_CVPR,qi2021offboard}. 
It is worth highlighting that the DETR-like paradigm has already surpassed CNN-based detectors and achieved state-of-the-art performance in the 2D object detection task. However, the DETR-like paradigm has not been fully explored in the context of 3D object detection. Our first attempt, STEMD, should be considered as a baseline for future, more powerful DETR-like models. There is still tremendous potential to be explored, as our framework has not incorporated several advanced techniques that have shown promising results in 2D DETR-like models~\cite{zhang2023dino,li2022dn,chen2023enhanced}.

Another limitation of our method is the lack of obvious improvement in detecting objects that are moving at high speeds. We attribute this result to the fact that we do not explicitly consider the motion of fast-moving objects in our spatial-temporal modeling. Even though the time interval between each frame is only 100 ms in the Waymo dataset, objects moving at high speeds can cover significant distances within this short time period. As a result, STGA-Net, our proposed model, fails to effectively capture the spatial-temporal dependencies between these fast-moving objects and their neighboring objects. In future research, it would be valuable to explore methods that can better handle the detection of fast-moving objects. Techniques such as motion estimation could be incorporated into the spatial-temporal modeling process to capture the dynamics of these objects more accurately. Additionally, investigating the use of higher frame rates or adaptive frame sampling strategies may help alleviate the issue of fast-moving objects being poorly represented in the temporal context.

In all, while our study has demonstrated the potential of spatial-temporal information for 3D object detection, there is still work to be done to improve the performance of our approach. By drawing inspiration from recent advancements in 2D DETR and addressing the challenges posed by fast-moving objects, future iterations of DETR-like models hold great promise for achieving even higher accuracy and robustness in multi-frame 3D object detection tasks.

\section{Conclusion}\label{sec:conclusion}
In this paper, we have presented STEMD, a novel end-to-end multi-frame 3D object detection framework based on the DETR-like paradigm. Our approach effectively models inter-object spatial interaction and complex temporal dependencies by introducing the spatial-temporal graph attention network, representing queries as nodes in a graph. Additionally, we improve the detection process by incorporating the detection results from the previous frame to enhance the query input of the decoder.
Furthermore, based on the characteristics of 3D detection tasks, we further incorporate the IoU regularization term in the loss function to reduce redundancy in bounding box predictions.
Through extensive experiments conducted on the two large-scale datasets, our framework has demonstrated superior performance in 3D object detection tasks.
The results validate the effectiveness of our proposed approach, showcasing the potential of modeling spatial-temporal relationships between objects in this domain.

% if have a single appendix:
%\appendix[Proof of the Zonklar Equations]
% or
%\appendix  % for no appendix heading
% do not use \section anymore after \appendix, only \section*
% is possibly needed

% use appendices with more than one appendix
% then use \section to start each appendix
% you must declare a \section before using any
% \subsection or using \label (\appendices by itself
% starts a section numbered zero.)
%

% \appendices
% \section{Notations}
% The notations used in this paper are summarized in Table~\ref{table:notations}.

% \setcounter{table}{0} % 将表格计数器重置为0

% you can choose not to have a title for an appendix
% if you want by leaving the argument blank
%\section{}
%Appendix two text goes here.

% use section* for acknowledgment
\ifCLASSOPTIONcompsoc
  % The Computer Society usually uses the plural form
%  \section*{Acknowledgments}
\else
  % regular IEEE prefers the singular form
%  \section*{Acknowledgment}
\fi

% The authors would like to thank...

% Can use something like this to put references on a page
% by themselves when using endfloat and the captionsoff option.
\ifCLASSOPTIONcaptionsoff
  \newpage
\fi

% trigger a \newpage just before the given reference
% number - used to balance the columns on the last page
% adjust value as needed - may need to be readjusted if
% the document is modified later
%\IEEEtriggeratref{8}
% The "triggered" command can be changed if desired:
%\IEEEtriggercmd{\enlargethispage{-5in}}

% references section

% can use a bibliography generated by BibTeX as a .bbl file
% BibTeX documentation can be easily obtained at:
% http://mirror.ctan.org/biblio/bibtex/contrib/doc/
% The IEEEtran BibTeX style support page is at:
% http://www.michaelshell.org/tex/ieeetran/bibtex/
\bibliographystyle{IEEEtran}
% argument is your BibTeX string definitions and bibliography database(s)
\bibliography{bib}

\end{document}

%\title{Spatial-Temporal Graph Attention Network for Multi-Frame 3D Object Detection}
\title{Spatial-Temporal Graph Enhanced DETR Towards Multi-Frame 3D Object Detection \\ (Supplementary Materials)}

%缩写可以是: STEMD, STETR(3D), or STEP3D

%
%
% author names and IEEE memberships
% note positions of commas and nonbreaking spaces ( ~ ) LaTeX will not break
% a structure at a ~ so this keeps an author's name from being broken across
% two lines.
% use \thanks{} to gain access to the first footnote area
% a separate \thanks must be used for each paragraph as LaTeX2e's \thanks
% was not built to handle multiple paragraphs
%
%
%\IEEEcompsocitemizethanks is a special \thanks that produces the bulleted
% lists the Computer Society journals use for "first footnote" author
% affiliations. Use \IEEEcompsocthanksitem which works much like \item
% for each affiliation group. When not in compsoc mode,
% \IEEEcompsocitemizethanks becomes like \thanks and
% \IEEEcompsocthanksitem becomes a line break with idention. This
% facilitates dual compilation, although admittedly the differences in the
% desired content of \author between the different types of papers makes a
% one-size-fits-all approach a daunting prospect. For instance, compsoc 
% journal papers have the author affiliations above the "Manuscript
% received ..."  text while in non-compsoc journals this is reversed. Sigh.

\author{
	Yifan Zhang, Zhiyu Zhu, Junhui Hou,~\IEEEmembership{Senior Member, IEEE},  and Dapeng Wu,~\IEEEmembership{Fellow, IEEE}
 \thanks{All authors are with the Department of Computer Science, City University of Hong Kong, Hong Kong SAR.}
}

% note the % following the last \IEEEmembership and also \thanks - 
% these prevent an unwanted space from occurring between the last author name
% and the end of the author line. i.e., if you had this:
% 
% \author{....lastname \thanks{...} \thanks{...} }
%                     ^------------^------------^----Do not want these spaces!
%
% a space would be appended to the last name and could cause every name on that
% line to be shifted left slightly. This is one of those "LaTeX things". For
% instance, "\textbf{A} \textbf{B}" will typeset as "A B" not "AB". To get
% "AB" then you have to do: "\textbf{A}\textbf{B}"
% \thanks is no different in this regard, so shield the last } of each \thanks
% that ends a line with a % and do not let a space in before the next \thanks.
% Spaces after \IEEEmembership other than the last one are OK (and needed) as
% you are supposed to have spaces between the names. For what it is worth,
% this is a minor point as most people would not even notice if the said evil
% space somehow managed to creep in.

% The paper headers
\markboth{}%
{Shell \MakeLowercase{\textit{et al.}}: Bare Demo of IEEEtran.cls for Computer Society Journals}
\maketitle

% To allow for easy dual compilation without having to reenter the
% abstract/keywords data, the \IEEEtitleabstractindextext text will
% not be used in maketitle, but will appear (i.e., to be "transported")
% here as \IEEEdisplaynontitleabstractindextext when the compsoc 
% or transmag modes are not selected <OR> if conference mode is selected 
% - because all conference papers position the abstract like regular
% papers do.
\IEEEdisplaynontitleabstractindextext
% \IEEEdisplaynontitleabstractindextext has no effect when using
% compsoc or transmag under a non-conference mode.

% For peer review papers, you can put extra information on the cover
% page as needed:
% \ifCLASSOPTIONpeerreview
% \begin{center} \bfseries EDICS Category: 3-BBND \end{center}
% \fi
%
% For peerreview papers, this IEEEtran command inserts a page break and
% creates the second title. It will be ignored for other modes.
\IEEEpeerreviewmaketitle
\appendices
\section{Complements to the Methodology}
\subsection{Notations}
For the sake of clarity and ease of reference, the main symbols are summarized in Table~\ref{table:notations}.

\renewcommand{\arraystretch}{1.1}
\begin{table}[htbp]
	\centering
	\caption{Main notations in this paper.}
	\label{table:notations}
	%	\begin{tabular}{>{\centering\arraybackslash}p{3cm} >{\centering\arraybackslash}p{6cm}} 
	\begin{tabular}{p{1.5cm} p{6cm}} 
		\toprule
		Notation~ & Definition  \\ 
		\hline
		$T$ & Length of point cloud sequence. \\
		$t$ & Frame index. \\
		$I_t$ & The point cloud at $t$-th frame. \\
		$X_t$ & Features output by CNN. \\
		$H_t$ & Out of ConvGRU at $t$-th frame. \\
		$X_t^{E}$ & Features output by encoder. \\
		$\mathcal{B}^E_t$ & Proposal generated by encoder. \\
		$N_p$ & Number of selected proposals. \\
		$N_q$ & Number of object queries. \\
		$\mathcal{Q}_t^0$ &Initial Object queries at $t$-th frame. \\
		$\mathcal{B}^{D}_{t}$ & Boxes predicted by decoder. \\
		$\mathcal{S}^{D}_{t}$ & Confidence scores predicted by decoder. \\
		$\mathcal{Q}^{D}_{t}$ & Query embedding output by decoder. \\
		$\mathcal{\hat{B}}^{D}_{t}$ & Selected boxes for graph construction. \\
		$\mathcal{\hat{Q}}^{D}_{t}$ & Selected queries as nodes in the graph. \\
		$N_g$ & Number of nodes in a graph. \\
		$G_t^{s}$ & Spatial graph at $t$-th frame. \\
		$V_t^s$ & The set of nodes in $G_t^{s}$. \\
		$E_t^s$ & The set of edges in $G_t^{s}$. \\
		$v^{s}_{t,i}$ & The $i$-th node in $G_t^{s}$. \\
		$d_s$ & Distance threshold used to determine the presence of edges between nodes. \\
		$\alpha_{ij}$ & Attention coefficient between $i$-th node and $j$-th node in spatial graph. \\
		$G^u_{t-1}$ & The source graph at $(t-1)$-th frame. \\
		$G^u_{t}$ & The target graph at $t$-th frame. \\
		$\beta_{ij}$ & Attention coefficients between $i$-th node in target graph and $j$-th node in source graph. \\
		TQR & Temporal Query Recollection. \\
		$\mathcal{Q}^{0}_t$ & The initial query embeddings for the decoder. \\
		$N_{\text{res}}$ & Number of queries input recollected from last frame. \\
		$\mathcal{R}_b$ & IoU regularization term. \\
		\bottomrule
	\end{tabular}
\end{table}
\renewcommand{\arraystretch}{1.0}

% \subsection{Temporal-Aware Transformer}
\subsection{ConvGRU-based Feature Enhancement for Encoder}
\label{sec:temporal_transformer} % Sequence input and process, revisiting DETR
% 关于 ConvGRU 的描述可以放到稍微靠后的位置, 要不就算了吧，还要定义很多符号，也不是很重要的样子
% 这里描述下 encoder, decoder, and conv-gru
% discretize
%First, we pass through the sparse voxel grids of each frame to a 3D sparse convolution network to obtain sparse feature maps, and further transform them into 2D BEV features $X_t$.
The BEV features extracted by the CNN backbone, along with corresponding positional embeddings, are sent to the encoder. This allows for the capture of local structures and contextual information in the BEV representation using a local self-attention mechanism, resulting in the generation of $X^E_t$. In addition to the standard encoder, we incorporate ConvGRU to capture both spatial and temporal information from sequential 2D BEV feature maps $\{X^E_t\}_{t=1}^{T}$, resulting in enhanced encoder features denoted as $H_t$.  
The ConvGRU model consists of two main components: the convolutional component and the GRU component~\cite{Ballas2015}. The convolutional component is able to extract spatial features from each input feature map, while the GRU component is responsible for modeling the temporal dependencies between the feature maps at different time steps. The output $H_t$ of ConvGRU at each time step $t$ is obtained by:
\begin{equation}
	\resizebox{1\linewidth}{!}{$
		\begin{aligned}
			R_t &= \sigma(\texttt{Conv}(H_{t-1},U_r) +\texttt{Conv}(X_t,W_r)+b_r), \\
			Z_t &= \sigma(\texttt{Conv}(H_{t-1},U_z) +\texttt{Conv}(X_t,W_z)+b_z), \\
			\tilde{H}_t &= \tanh(\texttt{Conv}(R_t \odot H_{t-1},U_h)  +\texttt{Conv}(Y_t,W_h)+b_h), \\
			H_t &= Z_t \odot H_{t-1}  +(1-Z_t)\odot\tilde{H}_t,
		\end{aligned}
		$}
\end{equation}
where $\texttt{Conv}(\cdot, \cdot)$ denotes the convolution operation, $X_t$ is the input feature map at time $t$, $\sigma(\cdot)$ represents the sigmoid activation function, $\odot$ denotes element-wise multiplication, $R_t$ and $Z_t$ are the reset gate and update gate vectors, respectively, $\tilde{H}_t$ is the candidate hidden state at time $t$.
In these equations, $H_{t-1}$ is the hidden state at time $t-1$, $U_r$, $W_r$, $U_z$, $W_z$, $U_h$, $W_h$ represent the GRU weights, and $b_r$, $b_z$, $b_h$ are the biases.
%\yfnote{The description about ConvGRU can be simplified or deleted.}

The enhanced feature $H_t$ is then utilized to generate initial object proposals $\mathcal{B}^E_t$ through a feed-forward network (FFN) head.
With the $N_q$ object queries $\mathcal{Q}_t$ initialized with boxes including selected top $N_p$ scored box proposals, we perform self-attention between queries $\mathcal{Q}_t$ and cross-attention between query $\mathcal{Q}_t$ and the enhanced encoder features $H_t$ to update the queries layer-by-layer~\cite{zhu2021deformable}. 
We implement the encoder and decoder following existing work~\cite{zhu2021deformable,zhang2023dino,zhu2022conquer}, and describe them in this section making this paper self-contained.

\section{Additional Experimental Results}
%It would be beneficial to show results on both LEVEL\_1 and LEVEL\_2 set in the supplementary material. This would help understand the behavior of the proposed method.
We present the results of the ablation study on both LEVEL\_1 and LEVEL\_2 sets in Sec.~\ref{sec:more_ablation}. These results provide deeper insights into the behavior and effectiveness of the proposed method. And in Sec.~\ref{sec:nuscenes_category}, we report the detailed performance for each class on the nuScenes dataset.
\subsection{Additional Ablation Studies}\label{sec:more_ablation}
\setlength{\tabcolsep}{6pt}
\begin{table}[htbp]
	\centering
	\caption{Contribution of each component in the proposed method. The results of mAP and mAPH on LEVEL\_1 and LEVEL\_2 are reported. ``S.I." refers to the sequential point cloud input. ``S.T." denotes spatial-temporal graph attention network. ``TQR" means the temporal query recollection strategy. ``IoU Reg" is the IoU regularization term in the loss function.}
	\label{table:ablation_components}
	\begin{tabular}{cccc|cccc} 
		\hline
		\multirow{2}{*}{S.I.} & \multirow{2}{*}{S.T.} & \multirow{2}{*}{TQR} & \multirow{2}{*}{IoU Reg.} & \multicolumn{2}{c}{Level\_1} & \multicolumn{2}{c}{Level\_2}  \\
		&   & & & mAP   & mAPH & mAP   & mAPH  \\ 
		\hline
		$\times$& $\times$& $\times$& $\times$ & 76.78 & 75.18 & 71.13 & 69.62 \\
		\checkmark & $\times$& $\times$& $\times$ & 77.27 & 75.77 & 71.63 & 70.10 \\
		\checkmark & \checkmark & $\times$ & $\times$ & 79.18 & 77.54 & 73.73 & 72.15 \\
		\checkmark & \checkmark & \checkmark & $\times$ & 79.81 & 78.18 & 74.39 & 72.87 \\
		\checkmark & \checkmark & \checkmark & \checkmark & 80.58 & 79.09 & 75.22 & 73.75 \\
		\hline
	\end{tabular}
\end{table}
\setlength{\tabcolsep}{6pt}

\setlength{\tabcolsep}{7pt}
\begin{table}[htbp]
	\centering
	\caption{
		The results comparison of our STEMD with the baseline using a point concatenation strategy under different frame length settings. We report the results of mAP/mAPH on both LEVEL\_1 and LEVEL\_2 of the Waymo validation set.}
	\label{table:ablation_sequential_processing}	
	\begin{tabular}{c|c|cccc} 
		\specialrule{1pt}{0pt}{0pt}
		\multirow{2}{*}{Frames} & \multirow{2}{*}{Method} & \multicolumn{2}{c}{Level\_1} & \multicolumn{2}{c}{Level\_2}  \\
		&    & mAP   & mAPH & mAP   & mAPH            \\ 
		\hline
		\multirow{3}{*}{2}      & Concat.   & 76.27 & 74.76   & 70.58 & 69.08    \\
		& Ours      & 77.50 & 75.94   & 71.91 & 70.43    \\
		& Improvement      & +1.23  & +1.18    & +1.33  & +1.35     \\
		\hline
		\multirow{3}{*}{4}      & Concat.   & 76.85 & 75.17   & 71.13 & 69.62    \\
		& Ours      & 80.58 & 79.09  & 75.22 & 73.75    \\
		& Improvement      & +3.73  & +3.92    & +4.09  & +4.13     \\
		\hline
		\multirow{3}{*}{8}      & Concat.   & 76.77 & 75.23   & 71.17 & 69.68    \\
		& Ours      & 81.44 & 79.80   & 75.92 & 74.30    \\
		& Improvement      & +4.66  & +4.56    & +4.75  & +4.62     \\
		\hline
		\multirow{3}{*}{16}     & Concat.   & 76.95 & 75.33   & 71.22 & 69.65    \\
		& Ours      & 81.56 & 79.99   & 76.14 & 74.56    \\
		& Improvement      & +4.61  & +4.66    & +4.92  & +4.91            \\
		\specialrule{1pt}{0pt}{0pt}
	\end{tabular}
\end{table}
\setlength{\tabcolsep}{6pt}

\noindent\textbf{Effect of Key Components.}
Table~\ref{table:ablation_components} presents the ablation study results, showcasing the contribution of each component in the proposed method. Starting with the baseline configuration without any of the proposed enhancements, the model achieves mAP/mAPH scores of 76.78\%/75.18\% and 71.13\%/69.62\% on LEVEL\_1 and LEVEL\_2, respectively.
Changing the processing of multi-frame point cloud input from direct concatenation to operating it as a sequence improves the performance to 77.27\%/75.77\% on LEVEL\_1 and 71.63\%/70.10\% on LEVEL\_2.
Adding the spatial-temporal graph attention network further enhances the scores to 79.18\%/77.54\% (LEVEL\_1) and 73.73\%/72.15\% (LEVEL\_2), demonstrating the effectiveness of capturing spatial-temporal dependencies between objects. Incorporating the temporal query recollection strategy boosts the scores to 79.81\%/78.18\% (LEVEL\_1) and 74.39\%/72.87\% (LEVEL\_2), suggesting that supplementing current frame queries with previous frame predictions significantly aids in handling challenging detection scenarios. Finally, adding the IoU regularization term in the loss function yields the best results, with mAP/mAPH scores reaching 80.58\%/79.09\% (LEVEL\_1) and 75.22\%/73.75\% (LEVEL\_2). This confirms that the regularization term effectively reduces redundancy and improves the precision of bounding box predictions. Overall, each component contributes positively to the model's performance, with the combination of all components leading to the highest detection accuracy.\\

\noindent\textbf{Effect of our Multi-frame Design.}
As shown in Table~\ref{table:ablation_sequential_processing}, we compare the performance of our multi-frame design, STEMD, against the baseline using a simple point cloud concatenation strategy under different frame length settings. The baseline shows limited ability to model long-term relations among frames, with performance slightly deteriorating as the number of concatenated frames increases from 8 to 16. In contrast, STEMD achieves consistent improvements as the frame length increases. Specifically, STEMD demonstrates 1.35\%, 4.13\%, 4.62\%, and 4.91\% higher mAPH than the multi-frame baseline using point concatenation with 2, 4, 8, and 16 frames, respectively. These results highlight the effectiveness of our method in leveraging long-term temporal dependencies across multiple frames, showcasing its superior performance in multi-frame 3D object detection tasks.

\setlength{\tabcolsep}{8pt}
\begin{table}[t]
	\centering
	\caption{Comparison of STGA-Net and conventional multi-head self-attention. We report the results of mAP/mAPH on both LEVEL\_1 and LEVEL\_2.}
	\label{table:ablation_attention}
	\begin{tabular}{c|cccc} 
		\specialrule{1pt}{0pt}{0pt}
		\multirow{2}{*}{Setting} & \multicolumn{2}{c}{Level\_1} & \multicolumn{2}{c}{Level\_2}  \\
		& mAP   & mAPH    & mAP   & mAPH            \\ 
		\hline
		Self-Attention  & 78.92 & 77.46 & 73.29 & 71.85  \\
		STGA-Net & 80.58 & 79.09 & 75.22 & 73.75  \\ 
		\hline
		Diff   & +1.66  & +1.63  & +1.93  & +1.90   \\
		\specialrule{1pt}{0pt}{0pt}
	\end{tabular}	
\end{table}
\setlength{\tabcolsep}{8pt}

\setlength{\tabcolsep}{5pt}
\begin{table}[htbp]
	\centering
	\caption{Effect of the number of nodes $N_{g}$ adopted in STGA-Net. We report the recall rate of nodes in the graph with respect to the ground-truth bounding boxes at IoU thresholds of 0.7 and 0.5. We also report the results of mAP/mAPH on both LEVEL\_1 and LEVEL\_2.}
	\label{table:ablation_graph_node_number}
	\begin{tabular}{c|cc|cccc} 
		\specialrule{1pt}{0pt}{0pt}
		\multirow{2}{*}{$N_{g}$} & \multirow{2}{*}{Recall@0.5} & \multirow{2}{*}{Recall@0.7} & \multicolumn{2}{c}{LEVEL\_1} & \multicolumn{2}{c}{LEVEL\_2}  \\
		&   &   & mAP   & mAPH     & mAP   & mAPH\\ 
		\hline
		1300   & 88.10  & 97.05  & 79.61 & 77.98    & 74.18 & 72.64     \\
		1000   & 87.67  & 97.03  & 80.23 & 78.60    & 74.72 & 73.65     \\
		800    & 87.09  & 96.94  & \textbf{80.58} & \textbf{79.09}    & \textbf{75.22} & \textbf{73.75}     \\
		500    & 85.74  & 96.56  & 79.78 & 78.30    & 74.63 & 73.47     \\
		300    & 84.63  & 96.09  & 79.74 & 78.21    & 74.46 & 72.98     \\
		\specialrule{1pt}{0pt}{0pt}
	\end{tabular}
\end{table}
\setlength{\tabcolsep}{2pt}

\setlength{\tabcolsep}{6pt}
\begin{table*}[t]
	\centering
	\caption{Comparison with state-of-the-art methods on the nuScenes validation set across various categories. The abbreviations 'C.V.,' 'Ped.,' 'M.C.,' 'B.C.,' 'T.C.,' and 'B.R.' stand for construction vehicle, pedestrian, motorcycle, bicycle, traffic cone, and barrier, respectively. The 'Frames' column denotes the number of keyframes.}
	\label{table:nuscenes_val_per_class}
	%	\scalebox{0.87}{
		\begin{tabular}{l|c|c|cc|cccccccccc} 
			\specialrule{1pt}{0pt}{0pt}
			Method        & Publication & Frames & mAP  & NDS  & Car  & Truck & Bus  & Trailer & C.V. & Ped. & M.C. & B.C. & T.C. & B.R.  \\ 
			\hline
			TransPillars~\cite{luo2023transpillars}  & WACV'23      & 3      & 52.3 & -    & 84.0  & 52.4  & 62.0   & 34.3    & 18.9 & 77.9 & 55.2 & 27.6 & 55.4 & 55.1  \\
			CenterPoint~\cite{yin2021center}   & CVPR'21      & 1      & 55.5 & 64.3 & 83.8 & 52.9  & 65.6 & 34.5    & 16.1 & 82.7 & 53.5 & 36.1 & 64.1 & 65.6  \\
			Focals Conv~\cite{chen2022focal}   & CVPR'22      & 1      & 61.2 & 68.1 & 86.6 & 60.2  & 72.3 & 40.8    & 20.1 & 86.2 & 61.3 & 45.6 & 70.2 & 69.3  \\
			TransFusion-L~\cite{bai2022transfusion} & CVPR'22      & 1      & 65.1 & 70.1 & 86.7 & 60.4  & \textbf{75.3} & 41.6    & 24.6 & 86.8 & 71.8 & 56.5 & 74.4 & 71.8  \\
            \hline
			Ours      & -           & 3      & \textbf{67.5} & \textbf{71.6} & \textbf{87.5} & \textbf{62.7}  & 75.2 & \textbf{42.5}    & \textbf{28.9} & \textbf{88.3} & \textbf{75.1} & \textbf{63.8} & \textbf{78.0} & \textbf{72.2}  \\
			\specialrule{1pt}{0pt}{0pt}
		\end{tabular}
		%	}
\end{table*}
\setlength{\tabcolsep}{5pt}

\noindent\textbf{Effect of the Spatial-temporal Graph Attention Network.}
To showcase the effectiveness of the spatial-temporal graph attention network (STGA-Net), we conducted experiments comparing it with conventional multi-head self-attention modules. As presented in Table~\ref{table:ablation_attention}, STGA-Net significantly outperforms the standard self-attention mechanism. Specifically, STGA-Net achieves an improvement of 1.66\% and 1.63\% in LEVEL\_1 mAP and mAPH, respectively, and 1.93\% and 1.90\% in LEVEL\_2 mAP and mAPH, respectively. These results indicate that the STGA-Net more effectively models the complex spatial-temporal dependencies between objects by dynamically and contextually capturing their relationships, in contrast to the densely applied self-attention mechanism that lacks this level of nuanced interaction.

\noindent\textbf{Hyper-parameters of Graph Node Selection.}
In our approach, we incorporate graph node selection to address the issue of overlapping bounding boxes and associated queries generated by the decoder, resulting in a more streamlined graph structure for downstream graph-based learning. It is crucial to determine the appropriate number of filtered bounding boxes, denoted as \(N_{g}\), since an excessively large value can introduce redundant bounding boxes that impede the effectiveness of graph-based learning, while setting \(N_{g}\) too small may result in a lower recall rate of queries with respect to the ground-truth boxes. By analyzing Table~\ref{table:ablation_graph_node_number}, we observe a significant decrease in the recall rate at the 0.5 IoU threshold, dropping from 97.05\% to 96.09\% when \(N_{g}\) is reduced from 1300 to 300. The best detection accuracy, achieving 80.58\% mAP and 79.09\% mAPH on LEVEL\_1 and 75.22\% mAP and 73.75\% mAPH on LEVEL\_2, is obtained with \(N_{g} = 800\), indicating that this value strikes a balance between recall and redundancy within the graph.\\

\setlength{\tabcolsep}{5pt}
\begin{table}[htbp]
	\centering
	\caption{Effect of the number of queries in decoder initialized by the box predictions in the last frame, i.e., $N_{\text{res}}$. We report the recall rate of all queries input for the decoder with respect to the ground-truth bounding boxes at IoU thresholds of 0.7 and 0.5. We also report the results of mAP/mAPH on both LEVEL\_1 and LEVEL\_2.}
	\label{table:ablation_recollection}
	\begin{tabular}{c|cc|cccc} 
		\specialrule{1pt}{0pt}{0pt}
		\multirow{2}{*}{$N_{\text{res}}$} & \multirow{2}{*}{Recall@0.7} & \multirow{2}{*}{Recall@0.5} & \multicolumn{2}{c}{L1} & \multicolumn{2}{c}{L2}  \\
		&   &   & mAP   & mAPH     & mAP   & mAPH\\ 
		\hline
		0          & 68.89      & 88.80      & 80.06 & 78.42 & 74.69 & 73.12  \\
		100        & 69.06      & 89.00      & 80.23 & 78.59 & 74.82 & 73.32  \\
		200        & 69.45      & 89.22      & 80.27 & 78.58 & 75.01 & 73.60  \\
		300        & 69.75      & 89.31      & \textbf{80.58} & \textbf{79.09} & 75.22 & \textbf{73.75}  \\
		500        & 69.76      & 89.36      & 80.22 & 78.58 & \textbf{75.25} & 73.75  \\
		800        & 69.81      & 89.39      & 80.06 & 78.59 & 75.11 & 73.74  \\
		\specialrule{1pt}{0pt}{0pt}
	\end{tabular}
\end{table}
\setlength{\tabcolsep}{5pt}

\noindent\textbf{Hyper-parameters of Temporal Query Recollection.}
As shown in Table~\ref{table:ablation_recollection}, incorporating additional \(N_{\text{res}}\) queries initialized with boxes predicted from the previous frame improves both the recall of all queries input to the decoder with respect to the ground-truth bounding boxes and the overall mAPH. Specifically, the recall increases from 68.89\% to 69.75\% at the 0.7 IoU threshold and from 88.80\% to 89.31\% at the 0.5 IoU threshold. The detection accuracy also improves, achieving the highest mAPH with 300 additional queries. However, further increasing \(N_{\text{res}}\) from 300 to 500 or 800 does not significantly enhance the recall or detection accuracy. This indicates that the collected queries from the last frame effectively supplement the encoder's output in the current timestamp as extra input queries for the decoder. It is also evident that setting an optimal \(N_{\text{res}}\) is crucial, as too many recollected queries could lead to unnecessary computational load without additional benefits.

\subsection{Performance Breakdown on nuScenes}\label{sec:nuscenes_category}
We further report the detailed results of STEMD for each category on the nuScenes dataset in Table~\ref{table:nuscenes_val_per_class}.
We can observe that STEMD outperforms the state-of-the-art method TransFusion in most categories, especially on the bicycle (+7.3\% AP), the construction vehicle (+4.3\% AP), and the traffic cone (+3.6\% AP).

% if have a single appendix:
%\appendix[Proof of the Zonklar Equations]
% or
%\appendix  % for no appendix heading
% do not use \section anymore after \appendix, only \section*
% is possibly needed

% use appendices with more than one appendix
% then use \section to start each appendix
% you must declare a \section before using any
% \subsection or using \label (\appendices by itself
% starts a section numbered zero.)
%

% \appendices
% \section{Notations}
% The notations used in this paper are summarized in Table~\ref{table:notations}.

% \setcounter{table}{0} % 将表格计数器重置为0

% you can choose not to have a title for an appendix
% if you want by leaving the argument blank
%\section{}
%Appendix two text goes here.

% use section* for acknowledgment
\ifCLASSOPTIONcompsoc
  % The Computer Society usually uses the plural form
%  \section*{Acknowledgments}
\else
  % regular IEEE prefers the singular form
%  \section*{Acknowledgment}
\fi

% The authors would like to thank...

% Can use something like this to put references on a page
% by themselves when using endfloat and the captionsoff option.
\ifCLASSOPTIONcaptionsoff
  \newpage
\fi

% trigger a \newpage just before the given reference
% number - used to balance the columns on the last page
% adjust value as needed - may need to be readjusted if
% the document is modified later
%\IEEEtriggeratref{8}
% The "triggered" command can be changed if desired:
%\IEEEtriggercmd{\enlargethispage{-5in}}

% references section

% can use a bibliography generated by BibTeX as a .bbl file
% BibTeX documentation can be easily obtained at:
% http://mirror.ctan.org/biblio/bibtex/contrib/doc/
% The IEEEtran BibTeX style support page is at:
% http://www.michaelshell.org/tex/ieeetran/bibtex/
\bibliographystyle{IEEEtran}
% argument is your BibTeX string definitions and bibliography database(s)
\bibliography{multi_frame_3DOD}